\documentclass[twoside,11pt]{article}

\usepackage{blindtext}

\usepackage[margin=1in]{geometry}
\usepackage{amsmath,amssymb,amsthm}
\usepackage{graphicx}
\usepackage{natbib}
\usepackage{hyperref}

\usepackage{hyperref}

\usepackage{amsmath}
\usepackage{amssymb}
\usepackage{mathtools}

\usepackage{algorithm}
\usepackage{algorithmic}

\usepackage{xcolor}

\newcommand{\mcA}{\mathcal{A}}
\newcommand{\mcR}{\mathcal{R}}
\newcommand{\mcAF}{\mathcal{A}_{\full}}
\newcommand{\mcF}{\mathcal{F}}
\newcommand{\mcP}{\mathcal{P}}
\newcommand{\E}{\mathbb{E}}

\newtheorem{theorem}{Theorem}

\newtheorem{lemma}[theorem]{Lemma}
\newtheorem{corollary}[theorem]{Corollary}
\newtheorem{remark}[theorem]{Remark}
\newtheorem{example}[theorem]{Example}
\newtheorem{assumption}{Assumption}[section]
\newtheorem{definition}{Definition}

\DeclareMathOperator*{\full}{full}

\DeclareMathOperator*{\conv}{Conv}

\DeclareMathOperator{\diam}{diam}
\DeclareMathOperator*{\regret}{Regret}

\DeclareMathOperator*{\dist}{dist}

\usepackage[titletoc]{appendix}
\usepackage{lipsum}

\usepackage{bbm}

\usepackage{enumitem}
\usepackage[normalem]{ulem}

\newcommand{\RR}{ \mathbb{R} }

\newcommand{\spaceo}{\hspace{2 mm}}
\newcommand{\setsep}{ \spaceo | \spaceo}

\newcommand{\Prob}[1]{\mathbb{P}\left( #1 \right)}

\newcommand{\Abs}[1]{\left| #1 \right|}
\newcommand{\Set}[1]{\left\{ #1 \right\}}
\newcommand{\Brack}[1]{\left( #1 \right)}

\newcommand{\inner}[2]{\left< #1 , #2 \right>}

\newcommand{\Expsubidx}[2]{ \mathbb{E}_{#1} #2}

\newcommand{\norm}[1]{\left\|#1\right\|}

\newcommand{\eps}{\epsilon}

\newlength{\dhatheight}

\begin{document}

\title{Representative Action Selection for Large Action Space Bandit Families}

\author{
Quan Zhou\\
National University of Singapore\footnote{
\texttt{quan.zhou@nus.edu.sg}}
\and
Mark Kozdoba\\
Technion\\
\and
Shie Mannor\\
Technion\\
}

\maketitle

\begin{abstract}
We study the problem of selecting a subset from a large action space shared by a family of bandits.
In many natural situations, while the nominal set of actions is large, actions are highly correlated: many yield similar rewards across environments, making it wasteful to maintain the full set. 
Our aim is to understand whether it is possible---and how---to select a smaller set of representative actions that performs nearly as well as the full action space. 

Our main contribution is a surprisingly simple algorithm: repeatedly sample a bandit instance at random, solve it, and collect the optimal action. 
This algorithm can significantly reduce the action space when such correlations are present, without the need to know a-priori the correlation structure.
We provide theoretical guarantees on the performance of the algorithm and demonstrate its practical effectiveness through empirical comparisons with Combinatorial Bandit, Meta Learning Bandit and Zooming baselines.
\end{abstract}


\section{Introduction}
\label{sec:intro}
In many decision-making problems, decisions must be made rapidly as the environment evolves. Although the pool of options is often large, many options perform similarly across most scenarios. Yet identifying the best option requires exploration that grows with the size of the pool, and by the time it is found, the environment may have already changed.
This raises a natural question:
Can we retain only a small set of options so that we can quickly find a good one, while performing almost as well as using all options?

This problem arises naturally in applications such as route planning \citep{feng2025sequential} and inventory management \citep{namir2022decision}, where evaluating or maintaining a large number of options is computationally or operationally infeasible.
For example, in a pharmacy inventory setting, each drug corresponds to an option, and each patient represents a different situation. While the number of available drugs may be large, many share similar ingredients and therefore tend to produce similar outcomes for a given patient. In such cases, it is desirable to retain only a small number of representative drugs without significantly degrading overall effectiveness.

\subsection{The Subset Selection Problem}
\label{sec:intro_reward_approx}

We formalize this problem in the setting of a family of multi-armed bandits \citep{lai1985asymptotically,lattimore2020bandit}, where each situation corresponds to a bandit instance drawn from a distribution over environments. 
Let 
$\mcF$ be a family of bandits which share an action space $\mathcal{A}_{\full}$, and let $\mcP$ be a probability distribution over $\mcF$.
For a bandit instance $\theta \in \mathcal{F}$, let $\mu_a = \mu_a(\theta)$ denote the expected reward of an action $a \in \mcAF$ given $\theta$. 
Each $\mu_a$ is then a random variable as $\theta$ is sampled from $\mcP$,
and the collection \( \{\mu_a\}_{a \in \mathcal{A}_{\full}} \) constitutes a stochastic process indexed by the action space.  

In this paper we consider the similarity structure on the actions $\mcAF$ that is induced by the correlations of the variables $\mu_a$. In this setting, the actions $a$ and $a'$ will be considered similar if the rewards $\mu_a$ and $\mu_a'$ are similar for most bandits sampled from $\mcP$:
We require the following increment condition:
\begin{equation}
\forall u > 0,\;\Pr\left[|\mu_a - \mu_{a'}| \geq u\right] \leq 2 \exp\left(-\frac{u^2}{2\|a - a'\|_2^2}\right). 
\label{equ:increment-define}
\end{equation}
Thus, \( \{\mu_a\}_{a \in \mathcal{A}_{\full}} \) is an (non-centered) sub-Gaussian process \citep{talagrand2014upper}. 
To illustrate our results, we first consider the simpler case of a centered Gaussian process, and then extend the analysis to the more general sub-Gaussian setting.

Now, suppose that the decision-maker has access to a certain subset of actions, $\mathcal{A} \subset \mathcal{A}_{\full}$, rather than to the full set $\mcAF$. 
For a bandit instance $\theta$, define the regret of being restricted to subset $\mathcal{A}$ as the gap between the best outcome achievable with the full action space and the best outcome within $\mathcal{A}$:
\begin{equation}
{\textnormal{Regret}(\theta) := \max_{a\in\mathcal{A}_{\full}} \mu_a(\theta) - \max_{a'\in\mathcal{A}} \;\mu_{a'}(\theta).}
\label{equ:regret-define}
\end{equation}
Our objective then is to identify a small subset $\mathcal{A}$ that minimizes the expected regret $\E _{\theta\sim \mcP}\left[\regret\right]$, where the expectation is taken over all possible bandit instances. 

\subsection{The Subset Selection Algorithm}
We now introduce the subset selection algorithm itself, Algorithm~\ref{alg:smallvertices}. Conceptually, the algorithm is simple: 
Sample a bandit instance $\theta$ from a distribution 
$\mcP$ and solve it using an oracle (or practically, any of the standard bandit algorithms, such as UCB or TS), to find the optimal action $a^* = a^*(\theta)$. Add the action to the set of representation actions $\mcA$. Repeat the process $K$ times. 

\begin{figure}[h]
\centering
\includegraphics[width=0.8\textwidth,height=0.3\textheight]{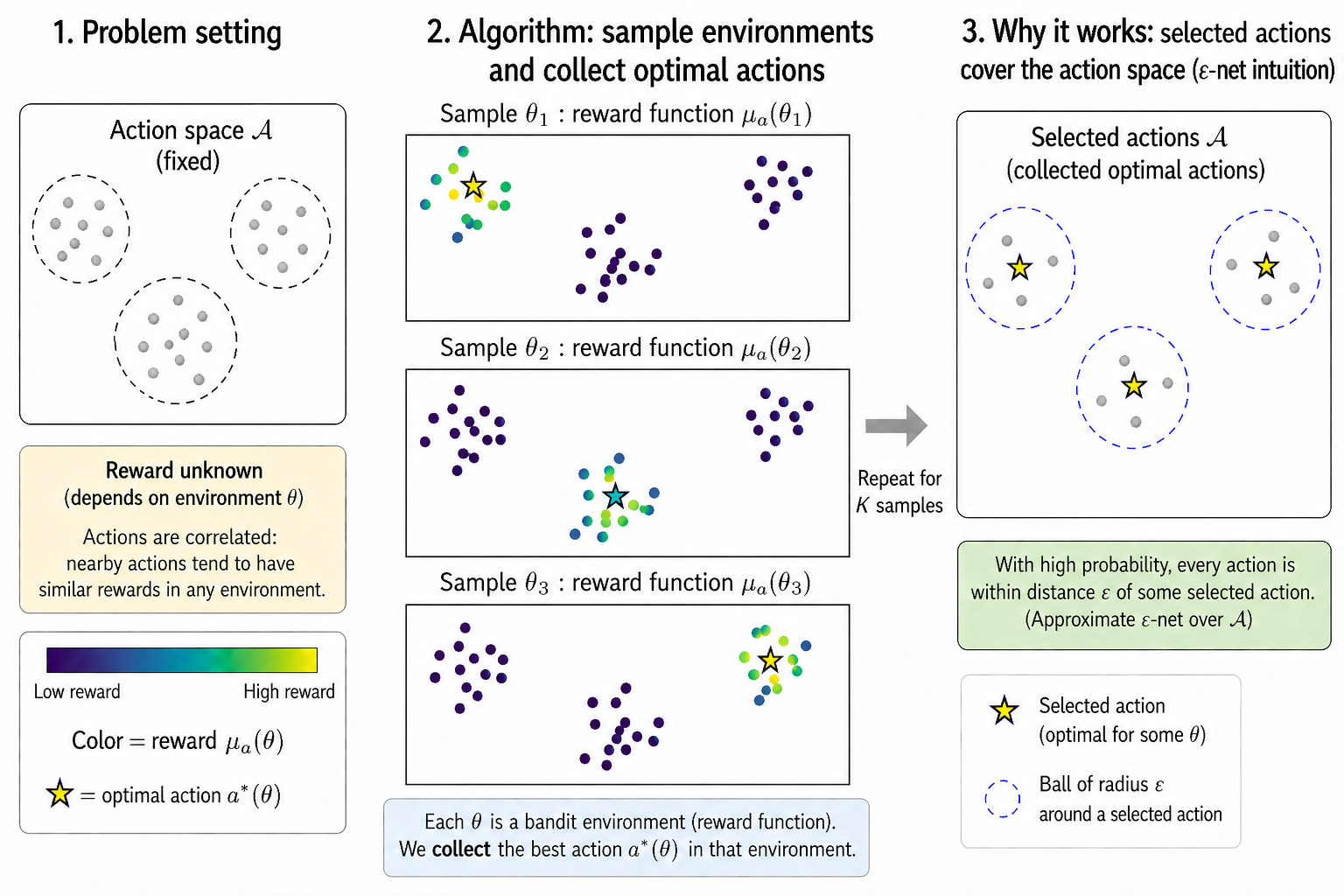}
\caption{\textbf{Conceptual overview of representative action selection.}
Consider a large action space exhibiting a clustering structure (left), where nearby actions have correlated rewards across environments.
Each environment $\theta$ induces a reward function over the same set of actions (middle; color indicates reward), and the optimal action varies across environments. 
The algorithm repeatedly samples environments and collects their optimal actions, forming a subset of representative actions (right).
Due to reward correlations, these selected actions effectively cover the action space, enabling low performance loss relative to the full action space.}
\label{fig:overview}
\end{figure}

Figure~\ref{fig:overview} illustrates the overall setup and the intuition behind the algorithm.

In the rest of the paper we show that the action subset generated by this simple algorithm will be diverse enough to appropriately cover $\mcAF$ by $\mcA$. Our bounds will depend on the metric structure of the set $\mcAF$ under the $L_2$ metric, and on the number of samples $K$. 

Note that Algorithm \ref{alg:smallvertices} does not require the knowledge of the metric, only the ability to sample bandit instances at random, and to solve them.

\begin{algorithm}[h]
\caption{Epsilon Net Algorithm}
\begin{algorithmic}[1]
\STATE \textbf{Input: } Action space $\mathcal{A}_{\full}$, Sample size $K$.
\STATE \textbf{Output: } A subset of actions $\mathcal{A}$.
\STATE $\mathcal{A}\gets\emptyset$
\FOR{$1,\dots, K$}
    \STATE Sample a bandit instance $\theta$ from a distribution $\mcP$ (unknown to the learner).
    \STATE Find optimal action $a^*(\theta)\!:=\!\arg\max_{a\in\mathcal{A}_{\full}} \mu_a(\theta)$.
    \STATE $\mathcal{A}\gets\mathcal{A}\cup\{a^*\}$.
\ENDFOR
\end{algorithmic}
\label{alg:smallvertices}
\end{algorithm}


\subsection{Motivations}
\label{sec:motivation}

Our objective in Equation~\eqref{equ:regret-define} may be motivated in several ways: 
First, as discussed above, in problems such as inventory management it may be of natural interest to find a representative subset of actions, due to the intrinsic difficulties associated with maintaining the full set of actions. 


Another possible reason would be to reduce the exploration phase when encountering a new bandit instance, especially when the time horizon $T$ is small.
Indeed, the classic Bayesian bandit regret, \citep{agrawal2012analysis}, generally scales as 
$\Omega(\sqrt{\Abs{\mcAF} T})$, and may become prohibitively large for  large $|\mcAF|$.
On the other hand, if a subset $\mathcal{A}$ is carefully chosen, the resulting Bayesian bandit regret can be significantly lower.
To see this, for a given $\mcA \subset \mcAF$ we can decompose the bandit regret as follows:
\begin{align}
&\textnormal{BayesianBanditRegret}:=\E \sum_{t=1}^T \left[\max_{a \in \mathcal{A}_{\full}} \mu_a(\theta) -  \mu_{A_t}(\theta)\right] \nonumber\\
&=\E \sum_{t=1}^T \left[\max_{a \in \mathcal{A}} \mu_a(\theta)\!-\!\mu_{A_t} +\max_{a \in \mathcal{A}_{\full}} \mu_a(\theta)\!-\!\max_{a \in \mathcal{A}} \mu_a(\theta)\right] \nonumber\\
&\leq  C\sqrt{|\mathcal{A}|\cdot T\log T} + T\cdot\E _{\theta}[\textnormal{Regret}]. \label{eq:base_bound_compare}
\end{align}
In the first equality, $A_t \in \mathcal{A}$ denotes the action selected in round $t$ by a policy such as Thompson Sampling (TS)~\citep{agrawal2012analysis}.
The expectation is taken over the randomness in the distribution of bandit instances and in actions selected by the policy.
The inequality \eqref{eq:base_bound_compare} then follows from the well-known regret bound for TS \citep{agrawal2012analysis,lattimore2020bandit}, and the definition of regret in \eqref{equ:regret-define}.
Thus, we can switch from a large regret 
$\sqrt{|\mathcal{A}_{\full}| \cdot T \log T}$, to a possibly much smaller one---$\sqrt{|\mathcal{A}| \cdot T \log T}$, although at a constant (per time step) price of $\E _{\theta}[\textnormal{Regret}]$. Practically, this means that by restricting to $\mcA$, the algorithm will spend much less time exploring similar options, and the above tradeoff may be especially worthwhile for small $T$, i.e., when rapid adaptation to a new bandit instance is desired. 

A further motivation is to reduce the computational burden that persists even in structured bandits; see Remark~\ref{rem:linear_and_gaussian_bandits} for a discussion of linear and Gaussian process (GP) bandits.

\subsection{Contributions}

Our main contributions are organized as follows:
\begin{itemize}[leftmargin=10pt]
\item \textbf{Problem Formalization and An Algorithm}  
Although we believe the subset selection problem among correlated actions is natural 
and of practical interest, to the best of our knowledge it has not been studied in the literature. In this paper we provide the first formalization of the problem, in the framework of the statistical similarities between actions. 
We also introduce a simple but effective subset selection algorithm, that does not require apriori  knowledge of 
the correlations.  

\item \textbf{Regret Analysis.}  
When the action sets exhibits sufficient correlation structure (ex. clustering or a low dimensional manifold structure), we provide upper and lower bounds on the expected performance loss of the algorithm, i.e. the regret \eqref{equ:regret-define} (Section \ref{sec:analysis}). The analysis is done under the 
standard assumption that the reward process $\{\mu_a\}_{a \in \mathcal{A}_{\full}}$ is Gaussian, \citep{srinivas2009gaussian,garnett2023bayesian} and shows the basic relations between the induced geometry of the action set and the expected rewards. 
Our arguments can also be applied virtually with no changes to the (non-centered) sub-Gaussian processes, as defined in \eqref{equ:increment-define}, and this extension is carried out later (Section \ref{sec:sub-gaussian-analysis}). 
An extension of these arguments to other types of tail behavior may be implemented using standard techniques.

\item \textbf{Generalization and Empirical Validation.}  
We extend the analysis to settings where outcome functions (or mean-reward functions) lie in a reproducing kernel Hilbert space (RKHS). 
We empirically evaluate the subset selection algorithm on a number of problems (Section \ref{sec:generalization}). 
We compare against three categories of baselines: (i) standard bandit algorithms (Thompson Sampling and UCB) applied to ``super-arms'', (ii) classic combinatorial bandit methods for action subset selection, and (iii) the Zooming algorithm \citep{kleinberg2008multi}, a state-of-the-art approach for optimization in large or continuous action spaces. We also benchmark against meta-learning bandit methods that leverage previous instances to construct a meta-prior and accelerate adaptation to new instances.
As expected, since these methods are not sensitive to the underlying geometry of the subsets, we show that their performance is weaker than that of our approach. 
\end{itemize}



\section{Related Work}

\textbf{Multi-Armed Bandits} \citep{lai1985asymptotically,auer2002finite} 
describe a sequential decision problem defined by a set of actions (arms), each deliver outcomes that are independently drawn from a fixed and unknown distribution. The decision-maker sequentially selects an action, observes its outcome, and aims to maximize cumulative outcomes over time. 
A decision-maker would follow a policy $\pi$ that chooses the next action based on the sequence of rounds and obtained outcomes. 
Popular policies include the upper confidence bound (UCB) \citep{auer2002finite}, Thompson sampling (TS) \citep{agrawal2012analysis}, and EXP3 \citep{auer2002nonstochastic} for adversarial settings.

\textbf{Connections to MAB:}
In the bandits problem, the objective is to maximize cumulative reward over a time horizon $T$ for a \emph{single, fixed bandit}. This is achieved by exploiting a single optimal action after exploration over the action space.
In contrast, our objective is to  select an action subset that likely contains near-optimal actions jointly for a \emph{family of bandits}. 


\vspace{0.1cm} \noindent \textbf{Optimal Action Identification}
focuses on identifying the action with the highest expected outcome in a MAB setting using as few samples as possible \citep{jamieson2014best,kaufmann2016complexity}.
Popular methods in fixed confidence setting include Action Elimination \citep{even2006action,karnin2013almost}, UCB, and LUCB, all of which achieve sample complexity within a $\log(|\mathcal{A}_{\full}|)$ factor of the optimum.
In fixed budget setting, there is successive halving \citep{karnin2013almost}, successive reject \citep{audibert2010best}.
Especially, \cite{gupta2021best} studies a setting where the rewards of different arms are correlated rather than independent and proposes a correlation-aware extension of the classical LUCB algorithm.

\vspace{0.1cm} \noindent \textbf{Linear Bandits} assumes that the expected outcome of each action depends \emph{linearly} on the inner product between an unknown parameter vector $\theta$ and the feature representation of an action $a \in \mathcal{A}_{\full}$ \citep{auer2002using,dani2008stochastic,rusmevichientong2010linearly}. 
This line of work assumes that action feature vectors are known \emph{a priori}, so the cardinality of the action space is not a limiting factor and may even be infinite. 
The bound scales as $O(n \sqrt{T})$, where $n$ is the dimension of the feature space.

\vspace{0.1cm} \noindent \textbf{GP Bandits} address the case where explicit feature representations are unavailable, but outcomes are assumed to be correlated according to a known kernel function $k(\cdot,\cdot)$ \citep{williams2006gaussian}. 
\citet{srinivas2009gaussian} model the reward function as a sample from a Gaussian process prior and propose the GP-UCB algorithm, which achieves cumulative regret $O\big(\sqrt{T \, \gamma_T \log |\mathcal{A}_{\full}|}\big)$ for finite action spaces, where $\gamma_T$ denotes the maximum information gain after $T$ rounds. 
The widespread adoption of this approach—both for finite action sets \citep{valko2013finite,li2022gaussian} and continuous domains \citep{chowdhury2017kernelized} highlights the practicality of assuming that action outcomes are correlated.

\vspace{0.1cm} \noindent
\textbf{Lipschitz Bandits} study bandit problems over continuous or very large action spaces, under the assumption that the reward function varies smoothly with respect to a metric on the action space \citep{kleinberg2008multi,slivkins2011contextual}.
Common approaches include the Zooming algorithm \citep{kleinberg2008multi}, which refines the exploration process according to the local geometry of near-optimal regions rather than relying on uniform discretization, achieving regret tied to the ``zooming dimension''; and Hierarchical Optimistic Optimization \citep{bubeck2011x}, which maintains a binary tree whose nodes correspond to measurable regions of the action space, explores by traversing the tree according to optimistic values, and progressively expands nodes to refine the partition in near-optimal regions.

This line of work is closely related to ours, as both deal with large action spaces endowed with a similarity structure, and the Zooming algorithm likewise exploits the notion of $\epsilon$-nets.
The key distinction is that Lipschitz bandits operate in an online setting on a single bandit instance, whereas our setting is offline subset selection and involves a family of bandit instances.
We compare empirically with the Zooming algorithm at Section~\ref{sec:online-bandits-exp}.

\vspace{0.1cm} \noindent \textbf{Meta Learning Bandits.}
This line of work comes from meta-learning \cite{baxter2000model,pontil2013excess}, also known as multi-task learning, which studies selecting a learning algorithm that performs well on tasks from a common environment (i.e., sampled from a prescribed distribution), relying on already completed tasks from the same environment. This setting has been extended to a class of bandit tasks \cite{deshmukh2017multi,cella2020meta}, where regret is analyzed across a sequence of related bandit instances. More recently, \cite{kveton2021meta,basu2021no} study how to improve exploration by meta-learning the prior distribution used in TS, alternating between estimating the prior from previous bandit instances and running TS on the current task using that estimate. 

This paper shares the same setup — bandit instances drawn from a distribution $\mathcal{P}$ — and the same goal of accelerating adaptation to new downstream tasks. The key distinction is that meta-learning bandits learn a shared prior online across tasks, whereas this paper performs an offline phase to compute a representative subset, which is then available to any downstream task without further meta-learning overhead.
We compare empirically with the MetaTS algorithm \citep{kveton2021meta} at Section~\ref{sec:online-bandits-exp}.

\vspace{0.1cm} \noindent 
\textbf{Latent Bandits \& Clustering Bandits}.
\cite{maillard2014latent} introduced the latent bandits framework, in which reward distributions are indexed by two sets — one for actions and one for types — and can be partitioned into a small number of clusters according to type. The decision-maker must choose an action for each type, where the sequence of types is generated according to some unknown stochastic process. \cite{hong2020latent} extends this framework and proposes algorithms based on UCB and Thompson sampling.
Several related frameworks share structural similarities. In clustering bandits \citep{gentile2014online}, users (or contexts) are drawn from a small number of hidden types, each sharing a similar reward model; the algorithm maintains at time $t$ an estimate of the context vector associated with each user type. This line is further extended to cluster both users and actions \cite{li2016collaborative}.
In routing bandits \citep{saber2021routine}, there is a finite but unknown set of bandit instances, and the learner chooses an arm for each bandit based only on past interactions. 
These works studies a related family of bandit problems, but focus on online regret minimization with discrete latent types. 
The underlying intuition that correlated actions can be collapsed to representatives, is the same as in this paper.
Essentially, the setting in this paper is a continuous-type latent bandit problem with a clustered action space. Our objective in \eqref{equ:regret-define} is offline subset construction using a sampling oracle, rather than online identification of the type. Due to the different objective (cumulative regret vs.\ one-shot subset regret), our techniques differ accordingly (UCB-on-types vs.\ $\epsilon$-nets).

\vspace{0.2cm}
Before discussing two related approaches that select a fixed-size subset $|\mathcal{A}| = K$--a choice often unclear in practice and requiring algorithm restarts--we highlight a key advantage of our method: it can adapt the subset cardinality on the fly.


\vspace{0.1cm} \noindent \textbf{Top-K Action Identification} aims to identify the $K$ actions with the highest expected outcomes using as few samples as possible \citep{kalyanakrishnan2012pac,gabillon2012best,kaufmann2016complexity,chen2017nearly}. This line of work assumes that all actions are independent and have distinct expected values, making its methods inapplicable to our framework.
If one were to apply these methods regardless, the most reasonable approach, in our view, would be to treat the family of bandit instances as a super-bandit, where each bandit instance corresponds to a round, and the expected payoff of an action in that round is given by $\mu_a$.
In this setting, top-$K$ identification would refer to selecting $K$ actions with the highest expected payoffs $\E [\mu_a]$.
In contrast, our framework considers that the expected outcome of each action, averaged over the distribution of bandits, may be the same---i.e., $\E_{\theta\sim\mathcal{P}} [\mu_a(\theta)] \!=\! 0$ for all $a \!\in\!\mathcal{A}_{\full}$.
Further, even if $\E [\mu_a]$ varies across actions, ignoring correlations can be fatal in our framework: 
\begin{example}\label{exm:top-k}
Consider three actions: $a_1 = [1, 0]$, $a_2 = [0.9, 0.1]$, and $a_3 = [-0.1, 1]$, and suppose bandits are sampled uniformly from $\theta_1 = [1, 0]$ and $\theta_2 = [0, 1]$. Then,
$$
\E \langle a_1, \theta \rangle = \E \langle a_2, \theta \rangle = 0.5,\quad \E \langle a_3, \theta \rangle = 0.45.
$$
So under the Best-$2$-Action perspective, $a_1$ and $a_2$ would be selected. However, this is suboptimal in our framework, since $\mu_{a_1}$ and $\mu_{a_2}$ are positively correlated:
\[
\E \max_{a\in\{a_1,a_2\}}\langle a, \theta \rangle=0.55,\quad \E \max_{a\in\{a_1,a_3\}}\langle a, \theta \rangle=1.
\]
Our algorithm, if run until it selects two distinct actions, would output $a_1$ and $a_3$---the true optimal.
\end{example}

\vspace{0.1cm} \noindent \textbf{Combinatorial Bandits}
considers that the decision maker selects $K$ of base arms from $\mathcal{A}_{\full}$ in each round, forming a super-arm $\mathcal{A}$, with $|\mathcal{A}|=K$ \citep{kveton2015tight}.
Popular methods include CUCB \citep{chen2016combinatorial}, CTS \citep{wang2018thompson}.
We include this line of work as an empirical baseline at Section~\ref{sec:combinatorial-exp}, though it is not directly applicable to our framework for two reasons: 
(1) It assumes that the expected outcome of a super-arm depends only on the expected outcomes of its individual base arms, or imposes a stricter monotonicity condition.
In our case, even though $\E [\mu_a] = c$ for all $a \in \mathcal{A}_{\full}$, super-arm expected outcomes $\E [\max_{a \in \mathcal{A}} \mu_a]$ can differ significantly due to correlations among actions.
(2) It assumes independence across base arms, whereas we explicitly model correlations. Ignoring these correlations misses the core challenge---an issue illustrated in Example~\ref{exm:top-k}.


\vspace{0.1cm} \noindent \textbf{Epsilon Nets} have two standard definitions.
The first, geometric definition \citep{vershynin2018high}, requires that radius-$\epsilon$ balls centered at net points cover the set. It relates to the covering number and extends to function classes, as in \citet{russo2013eluder}.
The second, measure-theoretic definition \citep{matousek2013lectures}, requires the net to intersect all subsets of sufficiently large measure.
The classic $\epsilon$-net algorithm by \citet{haussler1986epsilon} remains the simplest and most broadly applicable method. Later works aim to reduce net size \citep{pach2011tight,rabani2009explicit,mustafa2019computing} and address online settings \citep{bhore2024online}.

\vspace{0.1cm}
\noindent
\textbf{Expected supremum of a random process} for a given set $\mathcal{S}$ is the term $\E \left[\max_{a \in \mathcal{S}} \mu_a \right]$. It is an important topic in high-dimensional probability \citep{vershynin2018high} and asymptotic convex geometry \citep{rothvoss2021asymptotic}. The sharpest known bounds are due to \citep{talagrand2014upper}.
The expectation of regret defined in \eqref{equ:regret-define} is indeed the difference between the expected suprema of two random processes, $\{\mu_a\}_{a\in\mathcal{A}_{\full}}$ and $\{\mu_a\}_{a\in\mathcal{A}}$. This connection motivates our ideas and analysis around the concept of an $\epsilon$-net, which is a central tool for bounding the expected supremum.

\vspace{0.1cm} \noindent \textbf{Submodular Maximization.}
Submodular set function maximization under a cardinality constraint is a classical combinatorial optimization problem.
The greedy algorithm of \cite{nemhauser1978analysis} provides the foundational result: iteratively adding the element with the highest marginal gain yields a $1-1/e$ approximation guarantee for monotone submodular functions.
The continuous greedy algorithm of \cite{calinescu2011maximizing} relaxes the problem to the continuous domain via the multilinear extension, achieving the same $1-1/e$ guarantee for monotone functions under matroid constraints.
This line of work is relevant here because $f:2^{\mathcal{A}_{\text{full}}}\mapsto\mathbb{R}$, defined as $f(\mathcal{A})=\mathbb{E}_{\theta\sim\mathcal{P}}[\max_{a\in\mathcal{A}}\mu_a]$, is a monotone submodular set function, where we set $f(\emptyset)=-\infty$ to ensure monotonicity.

\section{Problem Setup and the Epsilon-Nets}
\label{sec:algorithm}
In this section, we introduce the formal problem setup and the technical tools used in the analysis. 

The optimal action in a bandit instance \( \theta \) is defined as  
\begin{equation}
a^*(\theta) := \arg\max_{a \in \mathcal{A}_{\full}} \mu_a(\theta).    
\end{equation}

Thus $a^*$ is a random variable taking values in the set 
$\mcAF$. We denote by $q$ the induced probability distribution of $a^*$ on $\mcAF$. 
Specifically, for any subset $r \subset \mcAF$, we have 
\begin{equation}
q(r):=\Prob{a^*(\theta)\in r} =
\Expsubidx{\theta \sim \mcP}{ \mathbbm{1}_{\{a^*(\theta)\in r\}} }.
\label{equ:q-measure}
\end{equation}
In particular, note that the samples $a^*$ in lines $5\&6$ of Algorithm \ref{alg:smallvertices} may be viewed as i.i.d draws from $q$.
The distribution $q$ will be named as the \emph{importance measure} $q$. 

\subsection{Gaussian Processes}
\label{sec:framework}

In our analysis we first consider the setting where the reward expectations  $\{\mu_a\}_{a \in \mathcal{A}_{\full}}$ form a Gaussian process, as a basis for developing the theory. We also extend our main bound to sub-Gaussian processes later in Section~\ref{sec:sub-gaussian-analysis}.

Let $\|\cdot\|_2$ denote the standard Euclidean norm on 
$\RR^n$, and for a set $S \subset \RR^n$ define the diameter of $S$ as $\mathrm{diam}(S) := \sup_{a,b \in S} \|a - b\|_2$.

To simplify the notation, we will consider the canonical representation of the Gaussian process:
That is, we assume that $\mcAF \subset \RR^n$ for some dimension $n>0$, and that 
\begin{equation}
{\mu_a(\theta) := \langle a, \theta \rangle, \quad \forall a\in\mathcal{A}_{\full},\theta \sim \mathcal{N}(0, I).}
\label{equ:mu-define}
\end{equation}
Here, $\inner{\cdot}{\cdot}$ is the standard inner product on $\RR^n$, and $\mathcal{N}(0, I)$ is the standard Gaussian distribution on $\RR^n$. Thus, in terms of the notation in Section \ref{sec:intro}, the family of bandits 
$\mcF$ may be identified as $\RR^n$, and the distribution $\mcP$ as the standard Gaussian. 
Clearly, the family of variables $\Set{\mu_a}_{a\in \mcAF}$ given by \eqref{equ:mu-define} is a Gaussian process. Conversely, we note that any finite dimensional Gaussian process can be represented as \eqref{equ:mu-define}, and any infinite dimensional Gaussian process\footnote{Under very mild technical assumptions} can be arbitrarily approximated by \eqref{equ:mu-define}, in a sufficiently high dimension $n$. See \citep[Chapter~7]{vershynin2018high}. 
Thus, we may assume the form \eqref{equ:mu-define} without loss of generality.

Moreover, crucially, the dimension $n$ will not be important in our bounds, and can be large.  On the other hand, however, it may be also of interest to note that when $n$ is fixed, the setup \eqref{equ:mu-define} is equivalent to the Bayesian \emph{linear bandit} setup, \citep{lattimore2020bandit} (although we solve a different problem). 
The relation with linear bandits will be discussed further in Remark \ref{rem:linear_and_gaussian_bandits}. 

We define the $L_2$ distance induced on the $\mcAF$ by the process to be the distance 
\begin{equation}
\label{eq:L2_distance_def}
        \norm{\mu_a - \mu_{a'}}^2_{L_2} := 
    \Expsubidx{\theta}{\Abs{\inner{a}{\theta} -\inner{a'}{\theta}}^2}.
\end{equation}
    
Note that under the definition \eqref{equ:mu-define}, the $L_2$ distance and the standard Euclidean distance in $\RR^n$ coincide, 
$\norm{\mu_a - \mu_{a'}}^2_{L_2} = \norm{a-a'}_2^2$ for all $a,a' \in \mcAF$.

\subsection{Epsilon Nets}
\label{sec:epsilon-nets}  
There will be two senses in which we would be interested in approximating $\mathcal{A}_{\full}$: First, for every 
$a\in \mcAF$ we would like to have $a' \in \mcA$ such that 
$a'$ is close to $a$ in the distance \eqref{eq:L2_distance_def} (or, equivalently, in $\norm{\cdot}_2$). Second, in order for the subset selection algorithm to able to discover the points $a'$, we need an additional, probabilistic notion of approximation.

The following two notions of an $\epsilon$-net, the \emph{geometric} and the \emph{measure theoretic} nets, 
formalize these approximations. We refer to 
 \citep[Chapter~4]{vershynin2018high}
 for additional details on geometric nets, and to 
 \citep[Chapter~10]{matousek2013lectures} for measure theoretic nets. 

\begin{definition}
A subset \( \mathcal{A} \subseteq \mathcal{A}_{\full} \) is called a \emph{geometric \(\epsilon\)-net} if, for all \( a \in \mathcal{A}_{\full} \), there exists \( a' \in \mathcal{A} \) such that  
\[\|a-a'\|_2 < \epsilon. \]      
\end{definition}

As discussed earlier, geometric $\epsilon$-nets will 
be used as approximations of the set $\mcAF$. While we'll assume that good approximating nets exist in the data, these nets will never need to be explicitly constructed. 
Instead, part of such representative points will be implicitly chosen by the subset selection algorithm. 

The $\epsilon$-\emph{covering number} of the set $\mcAF$,
denoted $N(\mcAF,\eps)$, is the size of the smallest 
$\epsilon$-net of $\mcAF$:
\begin{equation}
\label{eq:covering_number_definition}
    N(\mcAF,\eps) = \min \Set{ \Abs{\mcA'} \setsep 
    \mcA' \text{ is an $\epsilon$-net of $\mcAF$}}.
\end{equation}

Now, Let $\mathcal{R}$ be a finite partition of $\mcAF$ into disjoint clusters $r$, such that $\cup_{r\in\mathcal{R}}\; r = \mathcal{A}_{\full}$. Let  \( q \) be a probability distribution on $\mcAF$, which therefore assigns a weight $q(r)$ to each cluster \(r\in\mathcal{R}\).  
\begin{definition}
A subset \( \mathcal{A} \subseteq \mathcal{A}_{\full} \) is called a \emph{measure-theoretic \(\epsilon\)-net} with respect to measure \( q \) and partition $\mcR$, if, for any cluster \( r \in \mathcal{R} \), we have:  
 \[ r \cap \mathcal{A} \neq \emptyset \quad \text{whenever} \quad q(r) > \epsilon. \]      
\end{definition}

The measure theoretic $\epsilon$-net ensures that in a given partition $\mcR$, either $\mcA$ intersects a cluster $r$, or the cluster $r$ is small in terms of the distribution $q$. Roughly speaking, this will be used to show that if the subset selection algorithm doesn't find a cluster $r$ of actions, then the cluster's contribution to the total expected regret is small and can be ignored.

An important property of random sampling is that with high probability it will produce a measure theoretic $\epsilon$-net for \emph{any} fixed partition of the support of the distribution. Specifically, we have the following result:

\begin{lemma}\label{lem:measure-epsilon-net}
Fix $\epsilon>0$. Let  \( \mathcal{R} \) be a partition of $\mcAF$, and  let \( q \) be an importance measure given by \eqref{equ:q-measure}.
Let $\mathcal{A}$ be the output of Algorithm~\ref{alg:smallvertices} after $K$ samples. Then, with probability at least $1-\frac{1}{\epsilon} \exp(-K\epsilon)$, it holds that for any cluster $r\in\mathcal{R}$,
\[
r \cap  \mathcal{A} \neq \emptyset \quad \text{whenever} \quad q(r) > \epsilon.
\]
\end{lemma}

This result is a special case of Theorem~10.2.4 in \citep{matousek2013lectures}, which in turn follows the analysis in \citep{haussler1986epsilon}. The algorithm in 
\citep{haussler1986epsilon} is called the $\epsilon$-net Algorithm, and we have retained that name.  The full proof of the Lemma is given in Appendix \ref{app:easylemma-proof}.

\section{Regret Analysis}
\label{sec:analysis}
\begin{figure}
    \centering
    \includegraphics[width=0.4\linewidth]{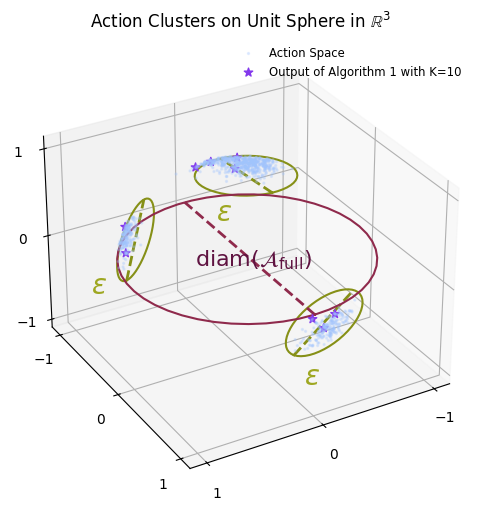}
    \caption{Illustration of Theorem~\ref{thm:alg-bound-upper}. The full action space (blue dots) lies on the unit sphere in $\mathbb{R}^3$ and can be partitioned into three clusters. The regret bound established in the theorem depends not on the diameter of the entire action space---indicated by the red dashed line---but rather on the diameters of the individual clusters, represented by the green dashed lines, each of which is significantly smaller.}
    \label{fig:Clusters_intro}
\end{figure}

In this section we introduce and discuss the regret bounds 
for subset selection Algorithm \ref{alg:smallvertices}.
Our main result is Theorem \ref{thm:alg-bound-upper}, 
which bounds the regret in terms of the covering numbers 
of the set $\mcAF$. Typically, sets which exhibit clustering, or sets with low dimensional manifold structure, would have low covering numbers, and we expect such sets to commonly occur in practical problems (the manifold hypothesis, \citep{bengio2013representation}). 
Roughly speaking, suppose that the set $\mcAF$ can be represented as a union of a few small clusters, as illustrated in Figure \ref{fig:Clusters_intro}. The main term in regret bound bound in Theorem \ref{thm:alg-bound-upper} is related to the complexity of each individual cluster, and typically would controlled by its diameter $\eps$. Additional terms involve a logarithmic dependence on the total number of clusters (times $\eps$), and an error term which accounts for the probability of Algorithm \ref{alg:smallvertices} to find the relevant clusters. This term decays exponentially with $K$. 

Perhaps the most central technical notion in our analysis is the notion of \emph{partition} $\mcR = \Set{r_{\ell}}_{\ell \leq m}$, such that $\mcAF = \cup_{l\leq m} r_{\ell}$, and we will typically be interested in partitions where each subset $r_{\ell}$ also have a relatively small diameter $\epsilon$.
There are two main ways in which such partitions may be present in $\mcAF$. First, some realistic datasets may be naturally clustered, as depicted in Figure \ref{fig:Clusters_intro}. Typically, in such situations there will be a characteristic scale $\eps$, such that there are few  clusters with diameter $\eps$, which are well separated. However, inside each cluster, on scales smaller then $\epsilon$, there may be no additional structure, i.e. the clusters themselves would look dense and high dimensional. In such situations, we still may benefit from a regret of order $\epsilon$. 
The second scenario is when the whole set $\mcAF$ is a low dimensional manifold, typically in a high dimensional space. In that case, for any scale $\epsilon$ we construct relatively small $\epsilon$-nets, which can be used to derive appropriate partitions.

Section~\ref{sec:ref_set_bounds} introduces bounds on the regret in terms of the partitions and their diameters. These are the bounds that one could obtain for a fixed a-priori \emph{known} partitions. In Section \ref{sec:alg-bound} we obtain bounds for the subset selected by Algorithm \ref{alg:smallvertices}, which finds (an approximation of) the reference points for the best partition available in $\mcAF$, without knowing this partition. Theorem \ref{thm:alg-bound-upper} provides bounds in terms of a general partition  given scale $\epsilon$, while in Theorem \ref{thm:alg-bound-upper-worst}, by taking $\epsilon$ to $0$ we provide worst case diminishing bounds. 

Finally, in Section~\ref{sec:sub-gaussian-analysis}, we relax the Gaussian process assumption and establish results in the general setting of a non-centered sub-Gaussian process.

\subsection{Partitions and Reference Set Bounds}
\label{sec:ref_set_bounds}

For a point $a\in \RR^n$ and scalar $\eps\geq 0$ denote 
by $B(a,\epsilon) := \Set{a' \in \RR^n \setsep \norm{a-a'}_2 \leq \epsilon}$ the closed Euclidean ball of radius $\epsilon$ around $a$. 


\begin{definition}[$\epsilon$-reference sets]\label{def:ref-subset}
Consider a partition \( \mathcal{R} := \{r_{\ell}\}_{\ell \leq m} \) of $\mcAF$. For an $\epsilon>0$, 
a set \( \mathcal{A}^{\epsilon} := \{a_1, \dots, a_m\} \) is an \emph{$\epsilon$-reference set} for $\mcR$ if 
for every $\ell \leq m$, 
\( r_{\ell} \subseteq B(a_{\ell}, \epsilon) \).
The point $a_{\ell}$ is called the \emph{reference point} of the subset $r_{\ell}$. 
\end{definition}

Note that the $\epsilon$-reference set is in particular a
(geometric) $\epsilon$-net of $\mcAF$. However, for our purposes it will be more convenient to discuss partitions, which we do with the notion of reference sets. 
Note also that given a partition $\mcR$, one can always find a reference set with $\epsilon = \max_{r\in \mcR} \diam(r)$.

\begin{theorem}[Regret bounds of $\epsilon$-reference subsets]\label{thm:geometric-bounds}
Let $\mcR$ be a partition of $\mcAF$ and let $\mathcal{A}^{\epsilon}$ be an $\epsilon$-reference subset for $\mcR$. There is an absolute constant \( C > 0 \), such that
\begin{equation*}
\E _{\theta}[\regret] \leq \max_{\ell\leq m}\E _{\theta}
\left[
\max_{a\in r_{\ell}} \Brack{\mu_a - \mu_{a_{\ell}}}
\right] + C\epsilon\sqrt{\log m}.
\end{equation*}

\end{theorem}

\textit{Proof sketch.}
For each cluster $\ell\leq m$, define a simple GP $\left\{Z_a\right\}_{a\in r_{\ell}}$ where $Z_a:=\mu_a-\mu_{a_{\ell}}$.  
Define a non-negative random variable $Y_{\ell}:=\sup_{a\in r_{\ell}} Z_a$. When $a^*(\theta)\in r_{\ell}$, regret is upper bounded by $Y_{\ell}$. 
Thus, for any bandit $\theta$, the regret is bounded between $\min_{\ell\leq m} Y_{\ell}$ and $\max_{\ell\leq m} Y_{\ell}$. 
Finally, the expectations $\E \left[\min_{\ell\leq m} Y_{\ell}\right]$ and $\E \left[\max_{\ell\leq m} Y_{\ell}\right]$ can be bounded via appropriate concentration arguments.
\hfill$\square$

We give the corresponding lower bound.
We define the outer radius induced by an action subset $\mathcal{S}$ as
\[
\Delta(\mathcal{S}):=\max_{a\in\mathcal{S}}\sup_{a'\in\mcAF}\|a-a'\|_2,
\]
which satisfies $\Delta(\mathcal{S})\le\diam(\mcAF)$.

\begin{theorem}[Lower bound of $\epsilon$-reference subsets]
\label{thm:geometric-bounds-lower}
Let $\mcR$ be a partition of $\mcAF$ and let $\mathcal{A}^{\epsilon}$ be an $\epsilon$-reference subset for $\mcR$. There is an absolute constant \( C > 0 \), such that
\begin{align*}
\E_{\theta}[\regret]
\ge
\max_{\ell \le m}\E\left[ \max_{a\in r_{\ell}} (\mu_a-\mu_{a_{\ell}}) \right]-C\;\Delta(\mathcal{A}^{\epsilon})\;\sqrt{\log m}.
\end{align*}
\end{theorem}
\textit{Proof sketch.}
Define $Y_{\ell,\ell'} := \sup_{a \in r_\ell \cup \{a_{\ell'}\}} (\mu_a - \mu_{a_{\ell'}})$. 
When $a^*(\theta)\in r_\ell$, we have an interesting property that
\[
\regret = \min_{\ell'} Y_{\ell,\ell'} \;=\; \min_{\ell'} \max_{\ell''} Y_{\ell'',\ell'}.
\]
Decomposing $Y_{\ell,\ell'}$ into its mean and deviation, the fluctuation term is controlled uniformly via Gaussian concentration, yielding a $\Delta(\mathcal{A}^{\epsilon})\sqrt{\log m^2}$ penalty. 
Finally, a min--max argument shows
\[
\min_{\ell'}\max_{\ell} \E Y_{\ell,\ell'} 
\;\ge\; 
\max_{\ell} \E\!\left[\max_{a\in r_\ell}(\mu_a-\mu_{a_\ell})\right],
\]
which gives the result.
\hfill$\square$

\subsection{Regret Bounds for the Algorithm}
\label{sec:alg-bound}

As discussed above, Theorem \ref{thm:alg-bound-upper} bounds the expected regret of the set $\mcA$ selected by Algorithm \ref{alg:smallvertices}.

To get an intuition about the argument, suppose 
that $\mcAF$ admits a good partition $\mcR$ 
(i.e. relatively few clusters, of small diameter $\eps$). 
We then show that for any fixed partition $\mcR$, 
samples in Algorithm \ref{alg:smallvertices} will hit most of the clusters of that partition, and can consequently be used as reference set points in Theorem \ref{thm:geometric-bounds}. Indeed, by Lemma \ref{lem:measure-epsilon-net}, 
either Algorithm \ref{alg:smallvertices} will hit the cluster, or the probability of that cluster under $q$ is small, in which case its contribution to regret will be small too. This gives rise to the additional error term in
Theorem \ref{thm:alg-bound-upper} compared to Theorem \ref{thm:geometric-bounds}, but this term decays exponentially with $K$.

\begin{theorem}\label{thm:alg-bound-upper} 
Let $\mcR$ be a partition of $\mcAF$ and set 
$\epsilon = \max_{r\in \mcR} diam(r)$.
Let $\mathcal{A}$ be the output of Algorithm~\ref{alg:smallvertices} after $K$ iterations.
Then 
\begin{equation*}
\begin{split}
\E_{\theta,\mathcal{A}}[\textnormal{Regret}] \leq &
\max_{\ell\leq m}\E _{\theta}
\left[\max_{a \in r_{\ell}} \Brack{\mu_a - \mu_{a_{\ell}}} \right]+ 
C\epsilon\sqrt{\log m} \\ & + \left(\E _{q}\left[(1-q(r))^{2K}\right]\cdot \E _{\theta}\left[\max_{a \in \mathcal{A}_{\full}} \mu^2_a\right]\right)^{1/2},
\end{split}
\end{equation*}
where $C>0$ is an absolute constant. 
\end{theorem}
Note that the expectation $\E _{\theta,\mathcal{A}}$
in this result is taken with respect to both the algorithm's randomness (i.e. the sampled set $\mathcal{A}$) and the distribution over $\theta$.

\textit{Proof sketch.}
If $r_{\ell} \cap \mathcal{A} \neq \emptyset$, choose $a_{\ell} \in r_{\ell} \cap \mathcal{A}$ as the representative point. If $r_{\ell} \cap \mathcal{A} = \emptyset$, choose an arbitrary point $a_{\ell} \in r_{\ell}$. The new set $\mathcal{A}' := \{a_{\ell}\}_{\ell \leq m}$ forms a reference subset.
Then, for each $\ell \leq m$, define a Gaussian process $\{Z_a\}_{a \in r_{\ell}}$, where $Z_a := \mu_a - \mu_{a_{\ell}}$, and let $Y_{\ell} := \sup_{a \in r_{\ell}} Z_a$. 
When $a^*(\theta) \in r_{\ell}$, we consider two cases. If $r_{\ell} \cap \mathcal{A} \neq \emptyset$, the regret is upper bounded by $Y_{\ell}$, and hence by $\max_{\ell \leq m} Y_{\ell}$. If $r_{\ell} \cap \mathcal{A} = \emptyset$, the regret is bounded by $\max_{a \in \mathcal{A}_{\full}} \mu_a$.
\hfill$\square$

In the following remarks we discuss the different error terms appearing in Theorem \ref{thm:alg-bound-upper}.

\begin{remark}[The Cluster Complexity Term]
The term $\E _{\theta}\left[\max_{a \in r_{\ell}} \mu_a -\mu_{a_{\ell}} \right]$ in Theorem~\ref{thm:alg-bound-upper} 
measures the size of the cluster, and is also known as the \emph{Gaussian width} of $r_{\ell}$ (see \citep{vershynin2018high}, Chapter 7). 
There are two standard ways in which we can bound this term:
First, for any finite $r_{\ell}$ we have 
\begin{equation}
\label{eq:gauss_width_bound}
 \E _{\theta}\left[\max_{a \in r_{\ell}} \mu_a - \mu_{a_{\ell}} \right] \leq \frac{\diam(r_{\ell})}{2} \sqrt{\log|r_{\ell}|} \leq 
 \frac{\eps}{2} \sqrt{\log \Abs{\mcAF}}, 
\end{equation}
thus yielding a scaling with the diameter and a square-root-logarithmic dependence on the size of the actions set. 
See Appendix \ref{app:bound-maxgaussian} additional details on \eqref{eq:gauss_width_bound}.
In addition, we have the bound 
\begin{equation}
\E _{\theta}\left[\max_{a \in r_{\ell}} \mu_a -\mu_{a_{\ell}}\right] \leq \diam(r_{\ell})\frac{\sqrt{n}}{2},    
\end{equation}
where $n$ is the ambient dimension of the action set, i.e.
$\mcAF \subset \RR^n$ (see \citep{vershynin2018high}, Chapter 7). This bound is independent of the cardinality of the action set and may be applied if $n$ is known to be small. However, as discussed in Section \ref{sec:framework}, we do not assume that $n$ is small. 
We use both bounds in Theorem \ref{thm:alg-bound-upper-worst} below. 
\end{remark}

\begin{remark}[Sampling Correction Term]
The sampling correction term 
 $\E _{q}\big[(1 - q(r))^{2K}\big]$ in Theorem~\ref{thm:alg-bound-upper} decays exponentially with $K$. Here we analyze this decay in more detail. Write 
\begin{flalign*}
\E _{q}\big[(1 - q(r))^{2K}\big] &= 
\sum_{r} q(r)(1 - q(r))^{2K} \\
&\le \sum_{r} q(r) e^{-2K q(r)},
\end{flalign*}
where we have used  $1 - x \leq e^{-x}$ for $x \in [0,1]$.
For each cluster $r \in \mathcal{R}$, the summand $q(r) e^{-2K q(r)}$ is small when $q(r)$ is small and decays rapidly when $q(r)$ is large.  
For any threshold $\tau \in (0,1)$, we split the sum into contributions from ``large'' and ``small'' probabilities:
\[
\sum_{r} q(r) e^{-2K q(r)} 
\le e^{-2K \tau} + \sum_{r:q(r)\leq \tau}q(r).
\]
Thus, to ensure $\E _{q}[(1 - q(r))^{2K}] \le \varepsilon$, it suffices to choose $\tau$ such that $\sum_{r:q(r)\leq \tau}q(r) \le \varepsilon/2$ and set
\[
K \ge \frac{1}{2\tau} \log\!\left(\frac{2}{\varepsilon}\right).
\]
\end{remark}

\subsubsection{Robustness to oracle perturbations}

The oracle step (Step~6) in Algorithm~\ref{alg:smallvertices} may appear restrictive. 
In practice, it can be relaxed by replacing the exact oracle with a best-arm identification (BAI) procedure. 
When $\mcA_{\full}$ is finite, classical BAI methods (e.g., \citep{audibert2010best}) can identify the optimal arm with arbitrarily high probability using a finite number of samples. 


\begin{corollary}[Approximate oracle via BAI]
\label{cor:approx-oracle}
Under the same conditions as Theorem~\ref{thm:alg-bound-upper}, suppose $\mcA_{\full}$ is large but finite, and the oracle step in Algorithm~\ref{alg:smallvertices} is replaced by a BAI procedure that identifies the optimal arm $a^*(\theta)$ with probability at least $1-\delta$. Then
\begin{equation*}
\begin{split}
\E_{\theta,\mcA}[\regret] \leq&
\max_{\ell\leq m}\E_{\theta}
\left[\max_{a \in r_{\ell}} (\mu_a - \mu_{a_{\ell}}) \right]
+ C\epsilon\sqrt{\log m} \\
&+ \left(
\E_{q}\!\left[(1-(1-\delta)\,q(r))^{2K}\right]
\cdot 
\E_{\theta}\!\left[\max_{a \in \mcA_{\full}} \mu_a^2\right]
\right)^{1/2},
\end{split}
\end{equation*}
where $C>0$ is an absolute constant.
\end{corollary}


\begin{proof}
The proof closely follows that of Theorem~\ref{thm:alg-bound-upper}. 
The only modification is in the upper bound on the probability of missing a cluster $r_\ell$ after $K$ samples. 
Let $\mcA$ denote the algorithm output, we have
\[
\Pr[\mcA \cap r_{\ell} = \emptyset]
\;\le\;
\left(1-(1-\delta)\, q(r_{\ell})\right)^K.
\]
Substituting this bound into the proof of Theorem~\ref{thm:alg-bound-upper} yields the desired results.
\end{proof}

\color{black}

\begin{remark}[Linear Bandits and Known Gaussian Process Regret Bounds]
\label{rem:linear_and_gaussian_bandits}    
As discussed in Section~\ref{sec:framework}, 
when dimension $n$ is fixed, we can view the problem setup 
\eqref{equ:mu-define} as that of Bayesian linear bandits. 
In Section~\ref{sec:motivation} we have mentioned two motivations for the action subset selections problem:
The intrinsic interest in reducing the size of the action set for purposes such inventory maintenance, and a possibility of faster learning in a new bandit instances, when the action set is reduced from $\mcAF$ to $\mcA$, see  
\eqref{eq:base_bound_compare}. 
While the first scenario holds for linear bandits as well, the second scenario may be weaker, since in linear bandits the regret may be bounded 
either as  $O(\sqrt{n \cdot \log \Abs{\mcAF}}\sqrt{T})$, \citep{russo2016information},  or as $O(n \sqrt{T})$, \citep{dani2008stochastic}.
Thus, in these situations  the regret depends at most only logarithmically on the size of $\Abs{\mcAF}$, \emph{if} the dimension $n$ is known to be small. However, as discussed earlier, we do not require the dimension to be small, and our bounds depend on the actual geometry of the set instead of relying on global constraints such as the dimension. Thus our bounds may remain small even when $n$ is arbitrarily large.

Similar considerations also apply more generally to existing regret bounds for Bayesian bandits with Gaussian Process reward distributions (a problem also known as Bayesian Optimization), \citep{srinivas2009gaussian, garnett2023bayesian}. While these bounds are applicable with infinite $\mcAF$, they involve global parameters of the model and assumptions on the kernel. For instance, for linear kernels the bound is $O(n\sqrt{T})$, while for RBF kernels one has $O(\sqrt{T \Brack{\log T}^{n+1} })$ - exponential in the underlying dimension, \citep{srinivas2009gaussian}.  Thus such bounds do not reflect the underlying clustering or non-linear low dimensionality of the data. 

Even for small $n$, each round incurs non-trivial optimization costs. Both UCB and TS require solving an inner maximization over $\mathcal{A}_{\text{full}}$, whose difficulty depends on the action space structure.
In linear bandits, UCB solves $\max_{a \in \mathcal{A}_{\text{full}}} \max_{\theta \in \mathcal{C}_t} \langle \theta, a \rangle$, which is NP-hard in general—even for convex $\mathcal{A}_{\text{full}}$~\citep{zhang2025tractable}—but trivial for small finite sets. TS similarly requires $\max_{a \in \mathcal{A}_{\text{full}}} \langle \tilde{\theta}_t, a \rangle$; while linear in $a$, this is costly for large finite sets, though near-sublinear costs $\tilde{O}(|\mathcal{A}_{\text{full}}|^{1-o(1)})$ are achievable under structural assumptions \citep{jun2017scalable,yang2022linear}. 
For GP bandits, the inner problem maximizes a non-concave acquisition function (e.g., GP-UCB), which is generally intractable for continuous domains. With finite $\mathcal{A}$ of size $|\mathcal{A}|$, the per-step cost is $\tilde{O}(|\mathcal{A}| h^2 + h^3)$ for batch size $h$~\citep{calandriello2022scaling}. Continuous domains typically require discretizations for approximate maximization~\citep{chowdhury2017kernelized,salgia2021domain}.


\begin{figure*}[!htbp]
\begin{minipage}[b]{0.45\textwidth}
\centering
\includegraphics[width=0.9\linewidth]{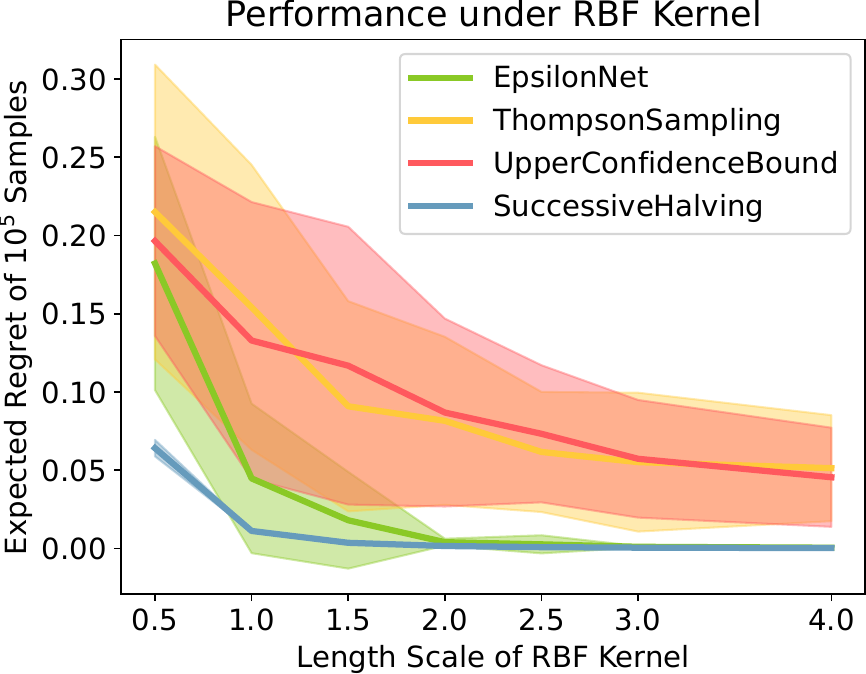}
\end{minipage}
\begin{minipage}[b]{0.51\textwidth}
\centering
\includegraphics[width=0.9\linewidth]{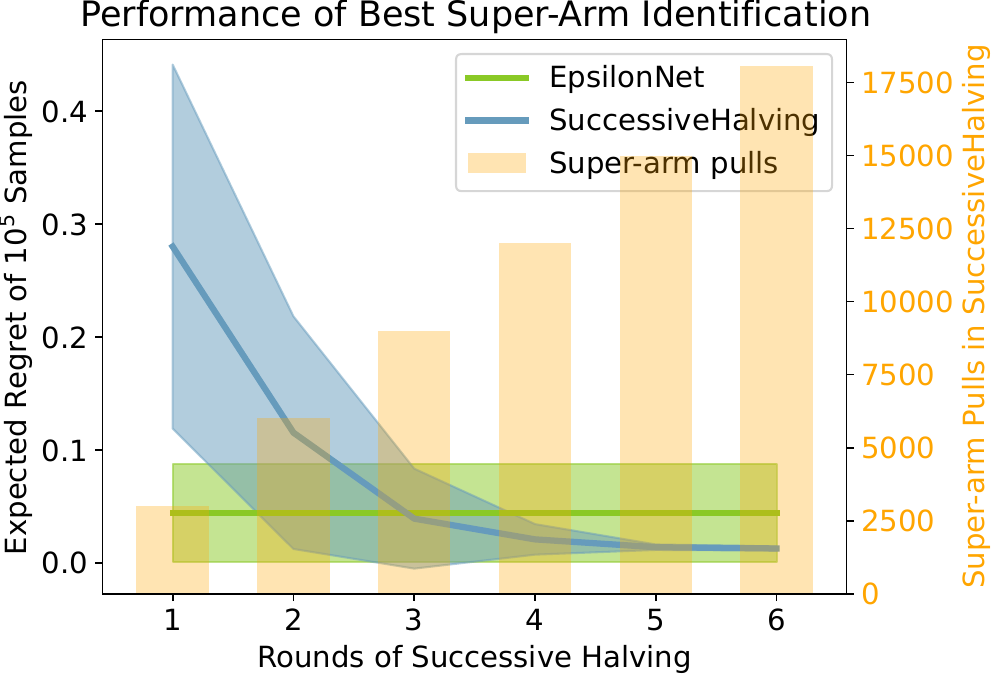}
\end{minipage}
\caption{
Comparison on solving~\eqref{equ:opt} by selecting $K=5$ actions from 15 grid points in $[0,2]$, with outcome functions $f(a)=\mu_a$ sampled via an RBF kernel at varying length-scales.  
\textbf{Left:} Expected regret over 50 repetitions: our method with exhaustive search (green) vs. super-arm TS (yellow), UCB (orange), and Successive Halving (SH; blue). TS/UCB run for 3,000 rounds; SH uses a budget of 37,000 pulls.  
\textbf{Right:} SH’s expected regret (blue) and cumulative super-arm pulls (yellow bars) per round vs. our method (green). SH requires nearly 10,000 pulls—about three times the number of super-arms—to match our performance.
}
\label{fig:RKHS_baselines}
\end{figure*}

\end{remark}

Finally, as discussed earlier, while Theorem \ref{thm:alg-bound-upper} is stated for a fixed partition, for any $\eps>0$ we can construct a partition derived from an $\eps$-net, and to obtain a corresponding regret bound. We carry this construction out in the following result. 
Recall the covering number $N(\mathcal{A}_{\full}, \epsilon)$ from \eqref{eq:covering_number_definition}.

\begin{theorem}\label{thm:alg-bound-upper-worst}  
Set $M := \diam(\mcAF)$ and fix $\eps>0$. 
Let \( \mathcal{A} \) be the output of Algorithm~\ref{alg:smallvertices}. Then, 
for 
\begin{equation}
\label{eq:worst_case_thm_K_condition}
 K \geq  c\cdot (M/\epsilon)^2\cdot N(\mathcal{A}_{\full},\epsilon),
\end{equation}
we have 
\begin{align}
\E _{\theta,\mathcal{A}}[\regret] \leq\; 
C\epsilon \min\{\sqrt{n},\sqrt{\log |\mathcal{A}_{\full}|}\} \nonumber + C'\epsilon\sqrt{\log N(\mathcal{A}_{\full},\epsilon)}, \nonumber 
\end{align}
where $C,C',c>0$ are absolute constants. 

In particular, by taking $\epsilon \to 0^+$, and taking $K$ to satisfy \eqref{eq:worst_case_thm_K_condition}, we have $\E _{\theta,\mathcal{A}}[\regret] \to 0$.
\end{theorem}

We now present a general lower bound that holds for any algorithm.
\begin{theorem}\label{thm:any-bounds-lower}
For any algorithm that outputs a subset $\mcA$ of cardinality $K$, the expected regret satisfies
\begin{align*}
\E_{\theta}[\regret]
\;\ge&\;
\max_{\ell \le m}
\E_{\theta}\!\left[
\max_{a \in r_{\ell}} \mu_a - \mu_{a_{\ell}}
\right]- C\;\Delta(\mcA)\sqrt{\log m + \log K},
\end{align*}
for some absolute constant $C>0$.
\end{theorem}

\textit{Proof sketch.}
The proof closely mimics that of Theorem~\ref{thm:geometric-bounds-lower}.
Define $Y_{\ell,i} := \sup_{a \in r_\ell \cup \{a_i\}} (\mu_a - \mu_{a_i})$. 
When $a^*(\theta)\in r_\ell$, $\regret = \min_i \max_{\ell} Y_{\ell,i}$. 
Decompose $Y_{\ell,i}$ into mean and fluctuation; the latter is uniformly controlled by Gaussian concentration, giving a $\Delta(\mcA)\sqrt{\log mK}$ term. 
Finally, a min--max argument lower bounds $\min_i \max_\ell \E Y_{\ell,i}$ by $\max_\ell \E[\max_{a\in r_\ell}(\mu_a-\mu_{a_\ell})]$.
\hfill$\square$

\begin{remark}
The leading term in Theorem~\ref{thm:any-bounds-lower},
\[
\max_{\ell \le m}
\E_{\theta}\!\left[
\max_{a \in r_{\ell}} (\mu_a - \mu_{a_{\ell}})
\right],
\]
coincides with the cluster-complexity term appearing in the upper bound (Theorem~\ref{thm:alg-bound-upper}) and captures the intrinsic difficulty of the problem. 
Moreover, when the output $\mcA$ is restricted to be a reference subset, so that $K=m$, this lower bound recovers the same qualitative scaling as the reference-subset lower bound in Theorem~\ref{thm:geometric-bounds-lower}, up to constants.
\end{remark}
\color{black}


\subsection{Regret Bound for Sub-Gaussian Process}
\label{sec:sub-gaussian-analysis}

The Gaussian process assumption can be relaxed to a sub-Gaussian one,
lifting the assumption in Equation~\eqref{equ:mu-define}.
We use Talagrand's generic chaining \citep{talagrand2014upper} to control
the supremum of such a process.

\begin{definition}
Let $T \subset \mathbb{R}^d$ be finite.  The \emph{$\gamma_2$ functional} is
\[
  \gamma_2(T)
  :=
  \inf_{(T_n)}\;
  \sup_{t \in T}
  \sum_{n \ge 0} 2^{n/2}\,\mathrm{dist}(t, T_n),
\]
where the infimum is over all \emph{admissible sequences} $(T_n)_{n\ge 0}$,
i.e.\ sequences of subsets of $T$ satisfying
$|T_0|=1$ and $|T_n|\le 2^{2^n}$ for all $n\ge 1$,
with $\mathrm{dist}(t,T_n):=\min_{s\in T_n}\|t-s\|_2$.
\end{definition}

Fix an admissible sequence $(T_n)$ achieving the infimum, and let
$\pi_n(t)\in T_n$ denote the nearest point in $T_n$ to $t$, such that $\mathrm{dist}(t, T_n)=\|t-\pi_n(t)\|_2$.
With $t_0$ the unique element of $T_0$, every $t\in T$ admits the
telescoping decomposition
\[
  X_t - X_{t_0}
  = \sum_{n\ge 1}\bigl(X_{\pi_n(t)}-X_{\pi_{n-1}(t)}\bigr),
\]
which rewrites the process over $T$ as a sum of increments at successive
resolution levels.
Since
\[
  \sup_{s,t\in T}|X_t-X_s|
  \le 2\sup_{t\in T}|X_t-X_{t_0}|
\]
by the triangle inequality, controlling the right-hand side via the
decomposition above suffices to bound the full supremum.

Using the generic chaining framework above, we derive two lemmas whose
combined role is to establish a regret bound for the sub-Gaussian case
in an analogous form to Theorem~\ref{thm:alg-bound-upper}.
The first controls the process increment within a single cluster $r_\ell$
in terms of the $\gamma_2$ functional $\gamma_2(r_\ell)$ and the diameter
$\mathrm{diam}(r_\ell)$, where $\gamma_2(r_\ell)$, also known as the
\emph{sub-Gaussian width} \citep{vershynin2018high}, is the analogue of
the cluster complexity term in Theorem~\ref{thm:alg-bound-upper}.
The second controls the squared process increment over the full action
space $\mathcal{A}_{\mathrm{full}}$, yielding the sub-Gaussian analogue
of the multiplier on the sampling correction term in
Theorem~\ref{thm:alg-bound-upper}.


\begin{lemma}\label{lem:sub-gaussian-tool}
Let $T \subset \mathbb{R}^d$ be finite.
Let the random process $\{X_t\}_{t\in T}$ satisfy the increment condition \eqref{equ:increment-define}.
Then, for every $u \ge 1$,
\[
\mathbb{P}\left(
\sup_{s,t \in T} |X_t - X_s|
\ge 16\,\gamma_2(T) + 8\,u\,\diam(T)
\right)
\le  2 e^{-u^2}.
\]
\end{lemma}
\textit{Proof sketch.}
We construct an admissible sequence of subsets $(T_n)$ for $T$.
Fix $u \ge 1$ and choose $\kappa$ such that
\[
2^{\kappa/2} \le u < 2^{(\kappa+1)/2}.
\]
The proof expands the increment $X_t - X_{t_0}$ along its telescoping chain induced by the projections $\pi_n(t)\in T_n$. But this decomposition is split into a coarse scale and a fine scale:
\[
X_t - X_{t_0}
=
\underbrace{(X_{\pi_\kappa(t)} - X_{t_0})}_{\text{coarse part}}
+
\underbrace{\sum_{n>\kappa} \big(X_{\pi_n(t)} - X_{\pi_{n-1}(t)}\big)}_{\text{fine part}}.
\]
The coarse part is controlled by a union bound over the finite net $T_\kappa$, while the fine part is handled by summing increments over all finer scales and applying union bounds over all pairs in $T_n \times T_{n-1}$. Combining both contributions yields the desired bound.
\hfill$\square$

\medskip

The next lemma is required because we do not assume a centered stochastic process.
Consequently, in the worst-case scenario where a cluster is missed by our algorithm and the optimal action lies within that cluster, the regret can be upper bounded by the looser quantity $\sup_{a,a'\in\mathcal{S}}|\mu_a-\mu_{a'}|$. The following lemma is then used to control this term. 

\begin{lemma}\label{lem:sub-Gaussian-deviation}
Let $\{X_t\}_{t\in T}$ be a sub-Gaussian process satisfying the increment condition \eqref{equ:increment-define}, then there exists an absolute constant $C>0$ that for any $u>0$
\[
\Pr\left(\sup_{s,t \in T} |X_t - X_s|\geq u\;\gamma_2(T)\right)\leq C\exp\left(-u^2/2\right).
\]
\end{lemma}
\textit{Proof sketch.}
This a simplified version of Lemma~\ref{lem:sub-gaussian-tool}.
We skip the coarse–fine decomposition and directly apply the standard generic chaining expansion of $X_t-X_{t_0}$ over all scales of an admissible sequence $(T_n)$. 
\hfill$\square$

\medskip

Using this functional, we establish a regret bound for our algorithm:

\begin{theorem}\label{thm:subGaussian-upper-bound} 
Let $\{\mu_a\}_{a\in\mathcal{A}_{\full}}$ be a sub-Gaussian process satisfying the increment condition in Equation~\eqref{equ:increment-define}. 
Let $\mathcal{A}$ be output of Algorithm~\ref{alg:smallvertices}.
Consider a partition \( \mathcal{R} := \{r_{\ell}\}_{\ell \leq m} \) of the full action space, where $m\ge 2$.
Then, for an absolute constant $C>0$, 
\begin{equation*}
\begin{split}
\E _{\theta,\mathcal{A}}[\textnormal{Regret}] 
\leq & 
16\,\max_{\ell\le m} \, \gamma_2(r_{\ell}) + 29\,\epsilon \sqrt{\log(m)}\\
& + C\cdot\left(\E _{q}\left[(1-q(r))^{2K}\right]\cdot (\gamma_2(\mathcal{A}_{\full}))^2\;\right)^{1/2}.
\end{split}
\end{equation*}
\end{theorem}

\textit{Proof sketch.}
The first part follows the same structure as the proof of
Theorem~\ref{thm:alg-bound-upper}.
If $r_{\ell} \cap \mathcal{A} \neq \emptyset$, choose
$a_{\ell} \in r_{\ell} \cap \mathcal{A}$ as the representative point;
otherwise, choose an arbitrary $a_{\ell} \in r_{\ell}$.
The set $\mathcal{A}' := \{a_{\ell}\}_{\ell \leq m}$ then forms an
$\epsilon$-reference subset.
For each $\ell \leq m$, define $Y_{\ell} := \sup_{a \in r_{\ell}}
(\mu_a - \mu_{a_{\ell}})$, and consider two cases when $a^*(\theta) \in r_{\ell}$.

If $r_{\ell} \cap \mathcal{A} \neq \emptyset$, the regret is bounded by
$\max_{\ell \leq m} Y_{\ell}$.
Applying Lemma~\ref{lem:sub-gaussian-tool} to each cluster and taking a
union bound over all $m$ clusters yields a concentration inequality for
$\max_{\ell \le m} Y_\ell$; integrating the tail then gives
\[
  \E\!\left[\max_{\ell \le m} Y_\ell\right]
  \le 16\,\max_{\ell \le m}\,\gamma_2(r_\ell) + 29\,\epsilon\sqrt{\log m}.
\]

If $r_{\ell} \cap \mathcal{A} = \emptyset$, the regret is bounded by
the looser quantity $\sup_{a,a'\in\mathcal{A}_{\mathrm{full}}}|\mu_a-\mu_{a'}|$,
since we do not assume a centered process.
Lemma~\ref{lem:sub-Gaussian-deviation} then yields
\[
\E[(\sup_{a,a'}|\mu_a - \mu_{a'}|)^2] \le C\,(\gamma_2(\mathcal{A}_{\mathrm{full}}))^2.
\]
Following the same
decomposition over sampled and missed clusters as in Theorem~\ref{thm:alg-bound-upper}, produces the final result.
\hfill$\square$

\section{Generalization and Empirical Validation}
\label{sec:generalization}

In general cases, the decision-maker may not have explicit access to the full structure of the action space, especially in high-dimensional settings. 
Instead, they are given a list of actions and can observe the expected outcomes of these actions through sampling.
We therefore treat the expected outcomes $\{\mu_a\}_{a \in \mathcal{A}_{\full}}$ as a family of random variables indexed by an abstract set.
In a single bandit, these expected outcomes define an \textbf{outcome function} \( f \) over the action space \( \mathcal{A}_{\full} \):  
$f(a) := \mu_a,\forall a \in \mathcal{A}_{\full}
$.

\paragraph{Applicability of our framework to RKHS: }
We first show that our framework applies when the outcome functions lie in a reproducing kernel Hilbert space (RKHS) \citep{williams2006gaussian}.
Consider a positive semidefinite kernel $k: \mathcal{A}_{\full} \times \mathcal{A}_{\full} \to \mathbb{R}$.
By Mercer's theorem, for a non-negative measure \( \mathbb{P} \) over \( \mathcal{A}_{\full} \), if the kernel satisfies  
\(
\int_{\mathcal{A}_{\full}\times\mathcal{A}_{\full}} k^2(a,a') d\mathbb{P}(a) d\mathbb{P}(a') < \infty,
\)
then it admits an eigenfunction expansion:
\(
k(a,a') = \sum\nolimits_{i\leq \infty} \lambda_i \phi_i(a) \phi_i(a'),
\)
where \( (\phi_i)_{i\leq\infty} \) are orthonormal eigenfunctions under \( \mathbb{P} \), and \( (\lambda_i)_{i\leq \infty} \) are non-negative eigenvalues.
Let outcome function\( f\) be of the form $f(\cdot)=\sum_{i=1}^N \alpha_i k(\cdot,a_i)$ for some integer $N\geq 1$, and a set of points $\{a_i\}_{i=1}^N\subset\mathcal{A}_{\full}$ and a weight vector $\alpha\in\mathbb{R}^N$. 
The function $f$ can be rewritten as 
\[
\mu_a = f(a) = \langle \mathbf{f}, 
\Phi(a) \rangle,
\]
where $\mathbf{f}$ is a vector of coefficients, and \( \Phi(a) \) usually refereed as a feature map, has entries \( \Phi_i(a) = \sqrt{\lambda_i}\phi_i(a) \).
The full action space is the one formed by the feature vectors $\Phi(a)$ for $a\in\mathcal{A}_{\full}$.

\paragraph{IID Actions: }
As an extreme case, suppose that $\{\mu_a\}_{a \in \mathcal{A}_{\full}}$ is a set of i.i.d. random variables.
This corresponds to the canonical process in which the action space is given by the orthonormal basis of $\mathbb{R}^n$, where $n = |\mathcal{A}_{\full}|$.
To see this, we associate each action $a$ with a unit vector $e_a$, which has a value of $1$ at the $a$th coordinate and $0$ elsewhere, and define the expected outcome as $\mu_a(\theta) := \langle e_a, \theta \rangle$.
With this construction, the collection $\{\mu_a\}_{a \in \mathcal{A}_{\full}}$ consists of mutually independent random variables.

\paragraph{Data Generation: }
Assuming the outcome function is sampled from a kernel implies action correlations. To study varying correlation levels, we use the RBF kernel  
\[
k(a, a') := \exp\!\left(-\frac{\|a - a'\|^2}{2l^2}\right),
\]  
where the length-scale \(l\) controls dependence among actions. Following~\citep{kanagawa2018gaussian} (Theorem~4.12), we sample functions from the RKHS by forming the kernel matrix \(\mathbf{K}_{a,a'} = k(a,a')\) over \(\mathcal{A}_{\full}\) and drawing \(f \sim \mathcal{N}(0, \mathbf{K})\). Although the kernel generates the outcomes, it is not revealed to the learner; thus, actions are treated as abstract indices rather than kernel inputs.

Returning to the objective stated in the Introduction, we seek a subset $\mathcal{A} \subseteq \mathcal{A}_{\full}$ of cardinality not more than $K$ that minimizes the expected regret defined in~\eqref{equ:regret-define}. Since $\E [\max_{a \in \mathcal{A}_{\full}} \mu_a]$ is constant with respect to $\mathcal{A}$, this is equivalent to solving:
\begin{equation}
\max_{\mathcal{A} \subseteq \mathcal{A}_{\full}} \E \!\left[\max_{a \in \mathcal{A}} \mu_a\right] \quad \text{subject to } |\mathcal{A}| \le K. \label{equ:opt}
\end{equation}
This problem is generally non-trivial: the objective involves an expectation over random variables with an unspecified (and potentially correlated) distribution, and the inner $\max$ operator induces non-smooth dependence on $\mathcal{A}$. Nevertheless, our method efficiently circumvents the combinatorial complexity inherent in~\eqref{equ:opt}.

\begin{figure*}[!htbp]
\begin{minipage}[b]{0.49\textwidth}
\centering
\includegraphics[width=0.82\linewidth]{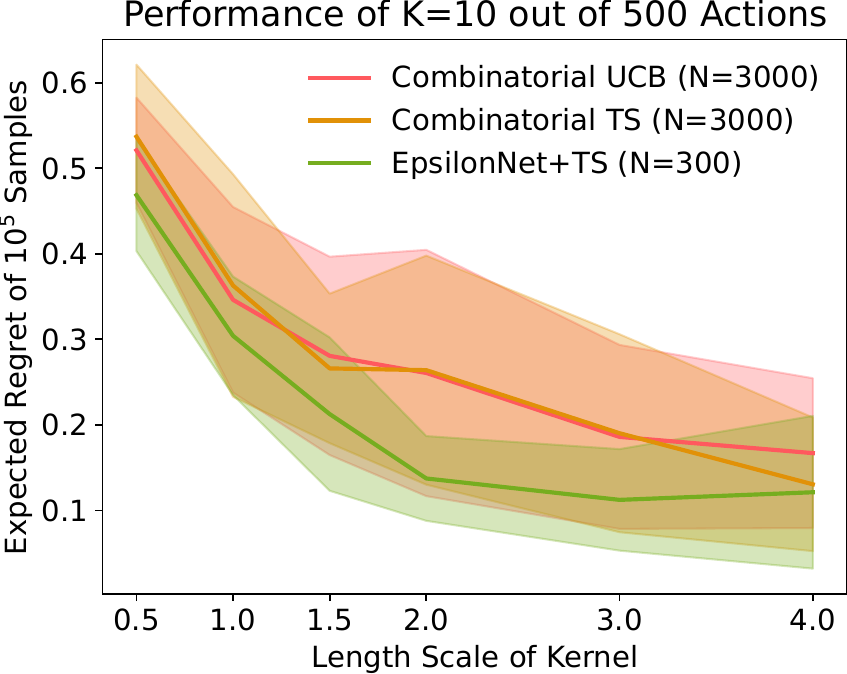}
\end{minipage}
\begin{minipage}[b]{0.55\textwidth}
\centering
\includegraphics[width=0.98\linewidth]{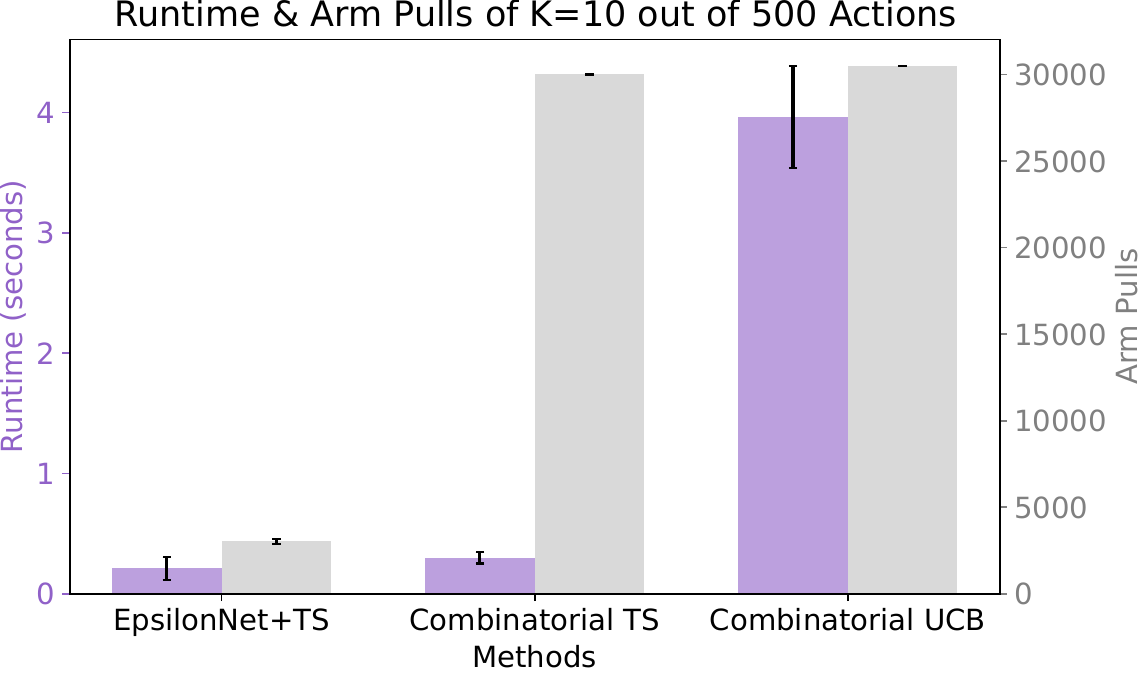}
\end{minipage}
\caption{Comparison of EpsilonNet+TS (green), CTS (brown), and CUCB (red) on solving~\eqref{equ:opt} with $K=10$ actions from 500 grid points in $[-5,5]$, using outcome functions $f(a)=\mu_a$ sampled via an RBF kernel at varying length-scales. In EpsilonNet+TS, the \texttt{argmax} oracle in Algorithm~\ref{alg:smallvertices} is replaced by TS; CTS and CUCB run for 3,000 rounds. All methods use a $\mathcal{N}(0,1)$ prior. \textbf{Left:} expected regret (30 repetitions). \textbf{Right:} runtime (purple, left axis) and arm pulls (gray, right axis).}
\label{fig:RKHS_combinatorial}
\end{figure*}

\subsection{Comparisons at the Super-Arm Level}
\label{sec:super-arm}

We compare our method against TS and UCB in terms of super-arm identification performance: we aim to identify near-optimal subsets of~\eqref{equ:opt} by selecting $K=5$ actions from an action space consisting of 15 grid points in $[0,2]$. Each 5-action tuple is treated as a super-arm. This yields $\binom{15}{5} = 3003$ super-arms---a moderate number that allows us to approximate the optimal super-arm using Successive Halving (SH), although SH becomes computationally infeasible for larger action spaces. Outcome functions are sampled via an RBF kernel at length scales $l \in \{0.5, 1, \dots, 4\}$.

In our approach, we run Algorithm~\ref{alg:smallvertices} until $K=5$ distinct actions are selected, using exhaustive search over the action space to compute the optimal action $a^*(\theta)$ for each bandit instance $\theta$.
TS and UCB operate in a bandit feedback setting: at each round (i.e., for a new $\theta$), a super-arm $\mathcal{A}$ is selected, the payoff $\max_{a \in \mathcal{A}} \mu_a(\theta)$ is observed, and the policy is updated. We run both methods for 3,000 rounds—matching the total number of super-arms ($\binom{15}{5} = 3003$).
The payoff of a super-arm $\mathcal{A}$ is defined as $\max_{a \in \mathcal{A}} \mu_a(\theta)$ to align with the objective in~\eqref{equ:opt}; thus, these methods directly optimize $\E [\max_{a \in \mathcal{A}} \mu_a]$. Since this payoff is unbounded, we use a $\mathcal{N}(0,1)$ prior for both TS and UCB.
For SH, we allocate a fixed budget of $N \log_2 N \approx 37{,}000$ arm pulls (following~\citep{karnin2013almost}), where $N = 3003$ is the number of super-arms. In each round (corresponding to a new $\theta$), the set of surviving super-arms is halved based on empirical payoffs, continuing until one remains or the budget is exhausted.
Finally, we estimate the expected regret of all methods over $10^5$ additional sampled outcome functions.

The left subplot of Figure~\ref{fig:RKHS_baselines} shows the expected regret of our method (green), TS (yellow), UCB (orange), and SH (blue). Solid curves and shaded regions denote the mean $\pm$ one standard deviation over $50$ repetitions. 
As observed, the expected regret decreases with increasing length-scale $l$.  
The length-scale $l$ controls the effective number of independent actions: for $|a - a'| \gg l$, $\mu_a$ and $\mu_{a'}$ become nearly uncorrelated. In the limit of large $l$, the outcomes $\{\mu_a\}_{a \in \mathcal{A}_{\text{full}}}$ behave like i.i.d. $\mathcal{N}(0,1)$ variables.


The right subplot of Figure~\ref{fig:RKHS_baselines} highlights SH’s impracticality. We ran both our method (green) and SH (blue) for 50 repetitions with fixed $l=1$, tracking the best super-arm found and total super-arm pulls after each SH round. The blue curves (with error shades) show SH’s expected regret (mean $\pm$ std.), while the yellow bars (right y-axis) indicate cumulative super-arm pulls per round.
SH matches our performance only after three rounds—but at roughly three times the total number of super-arms.



\subsection{Oracle-Free Comparisons with CTS and CUCB}
\label{sec:combinatorial-exp}

While Algorithm~\ref{alg:smallvertices} relies on an oracle \texttt{argmax} step, this is often infeasible in practice. We replace it with a simple TS rule, yielding a fully implementable, oracle-free variant (EpsilonNet+TS). We compare this method against combinatorial TS (CTS) and combinatorial UCB (CUCB)---two practical approaches for choosing an action subset with $|\mathcal{A}| = K$---under the same setup as in Section~\ref{sec:super-arm}, but now selecting $K=10$ actions from an action space consisting of 500 grid points over $[-5,5]$.


Our approach runs Algorithm~\ref{alg:smallvertices} until 10 distinct actions are identified. The oracle \texttt{argmax} step is approximated by running TS for 300 rounds; the action chosen in the final round serves as an approximation of $a^*(\theta)$. 
CTS and CUCB operate in a semi-bandit setting for 3000 rounds. In each round $t$, an instance $\theta_t$ is sampled, an action subset $\mathcal{A}_t$ of cardinality $K=10$ is selected, and rewards $\mu_a(\theta_t)$ are observed for all $a \in \mathcal{A}_t$. The subset selected in the final round is taken as the algorithm's output.
All methods use a $\mathcal{N}(0,1)$ prior, consistent with the ground truth.




In Figure~\ref{fig:RKHS_combinatorial}, the left subplot shows the expected regret (mean $\pm$ standard deviation over 30 repetitions) for EpsilonNet+TS (green), CTS (brown), and CUCB (red).  
As observed, the poor performance of CTS and CUCB aligns with discussions in related work.  
The right subplot reports runtime and arm pulls across 30 repetitions with fixed $l=1$: the left axis (purple) shows average runtime $\pm$ one standard deviation, while the right axis (gray) shows total base arm pulls.
EpsilonNet+TS uses exactly 3,000 arm pulls (300 TS rounds each $\times$  10 base arms). CTS incurs comparable runtime but requires far more pulls (3,000 CTS rounds $\times$ 10 base arm pulls). CUCB is significantly slower and incurs additional pulls due to its mandatory initialization.

\subsection{Online Evaluation on Downstream Bandit Instances}
\label{sec:online-bandits-exp}

We evaluate how effectively our method exploits meta-training information when faced with a new, unseen bandit instance $\theta$, i.e., a new outcome function $f(a)=\mu_a(\theta)$, for $a\in\mathcal{A}_{\full}$. 
The downstream bandit instance proceeds for $T$ rounds: at each round $t$, the decision-maker selects one action $a$ from a given action set and observes its true mean reward $\mu_a(\theta)$. 
We set $T = 500$, $K = 10$, and use $N = 500$ grid points on $[-5, 5]$ as the full action space with an RBF kernel of length scale $\ell = 1.0$. Results are averaged over $30$ repetitions, each evaluated on $100$ fresh test bandits instances drawn from the same GP prior.

We compare three methods on the downstream instances:

\begin{itemize}
\item \textbf{EpsilonNet+TS.} 
After meta-training on $K$ bandits instance (the oracle \texttt{argmax} step in Algorithm~\ref{alg:smallvertices} replaced by TS for $300$ rounds), the algorithm restricts the action space to the $K$-action representative subset $\mathcal{A}$.
In the downstream instances, TS policy is then run over the restricted subset $\mathcal{A}$ for $T$ rounds. We use $\mathcal{N}(0,1)$ prior for both TS rules.

\item \textbf{Zooming} \citep{kleinberg2008multi}. A meta-training-free baseline that treats the full action space as a Lipschitz metric space. Starting from a single seed arm at the center of the action space, the algorithm maintains an active set of arms and expands it by activating any arm whose nearest active neighbor's confidence ball no longer covers it. UCB is played over the active set. No meta-training data is used; all adaptation is purely online.

\item \textbf{MetaTS} \citep{kveton2021meta}. A meta-learning bandit baseline that uses the same $K$ meta-training instances, i.e, the true mean reward $\mu_a(\theta)$ for all $a\in\mathcal{A}_{\full}$ and for all $K$ instances, to construct an empirical Bayes prior. 
In the downstream instances, TS is run over the full action space, with the learned prior replacing the uninformative default. This baseline tests whether a warm-started prior over the full space can compete with explicit action-space reduction.
\end{itemize}

\begin{figure}
    \centering
    \includegraphics[width=\linewidth]{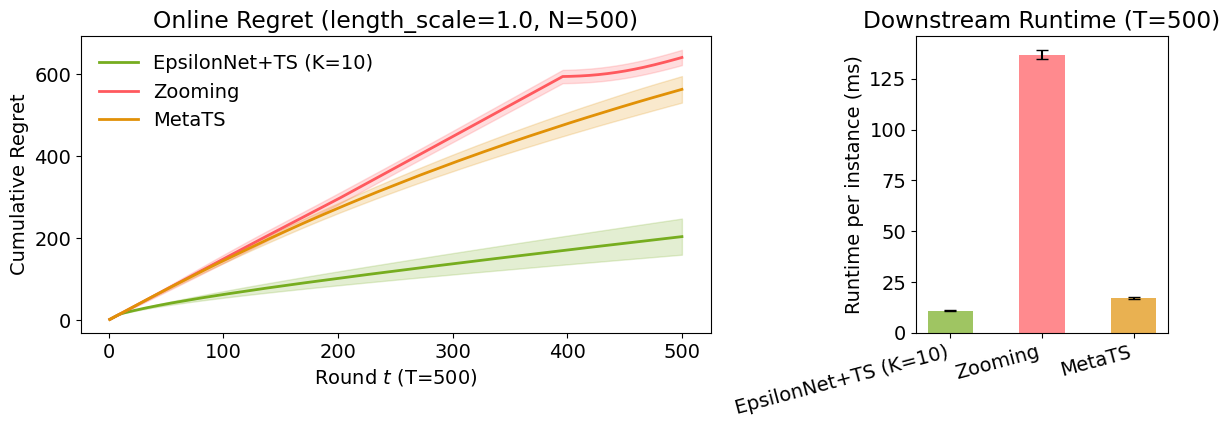}
\caption{Downstream bandit performance comparison with action space being $N=500$ grid points in $[-5,5]$, and outcome functions $f(a)=\mu_a$ sampled via an RBF kernel at length-scale $\ell=1$. Left panel: Cumulative regret in \eqref{equ:cumulative-regret} over $T=500$ rounds as a function of time, with shading indicating $\pm 1$ standard deviation over $30$ repetitions. Right panel: Mean runtime per downstream instance. EpsilonNet+TS achieves superior regret-runtime tradeoff by exploiting the meta-trained action-space reduction from $N=500$ to $K=10$ arms, enabling both faster decision-making and lower cumulative regret compared to meta-training-free (Zooming; \cite{kleinberg2008multi}) and full-space meta-learning (MetaTS; \cite{kveton2021meta}) baselines.}
    \label{fig:downstream-instance}
\end{figure}

In Figure~\ref{fig:downstream-instance}, we report two metrics: cumulative regret and runtime.

\paragraph{Cumulative regret (left panel).} The left panel shows cumulative regret, defined as below, which is a commonly used metric in the bandits literature \citep{lattimore2020bandit}:
\begin{equation}
\text{CumulativeRegret}(t):=     \max_{a\in\mathcal{A}_{\full}} \mu_a(\theta) - \sum_{s=1}^t  \mu_{a_s}(\theta),
    \label{equ:cumulative-regret}
\end{equation}
where $a_s$ is the action chosen at round $s$,
as a function of round $t$, with shading indicating $\pm 1$ standard deviation across meta-training repetitions. Our method ``EpsilonNet+TS'' achieves the lowest cumulative regret, with the gap over baselines widening in the early rounds where the smaller action space allows rapid concentration of probability mass on the best arm. MetaTS benefits from a warm prior but must still explore $N = 500$ arms, leading to higher regret than ours particularly when $T$ is small relative to $N$. 
Zooming starts with a single active arm and activates neighbors only as confidence radii shrink, so its regret initially grows faster; it eventually converges as the active set covers the high-reward region, but its final cumulative regret remains comparable to MetaTS.
The contrast is expected because we set the horizon of the downstream bandit instance equal to the cardinality of the full action space ($N=T=500$).

\paragraph{Runtime (right panel).} The right panel reports the mean runtime to complete $T = 500$ rounds of the downstream bandit instance. EpsilonNet+TS is the fastest, as its inner loop only samples from a $K$-dimensional posterior. MetaTS is moderately more expensive due to posterior updates over all $N=500$ arms. Zooming is the most expensive, because at each round it checks the coverage condition for every inactive arm before computing UCB over the active set. 

Taken together, EpsilonNet achieves the best regret--runtime trade-off: the meta-training phase pays a one-time cost to compress the action space, after which each downstream decision is both faster and more accurate.

\section{Conclusions and Future Work}
We proposed a framework for selecting a small representative subset of actions from a very large shared action space in a family of bandit problems. Our simple subset-selection algorithm exploits reward correlations to identify representative actions with controllable performance loss.
Natural directions for future work include developing data-dependent stopping rules for choosing the subset size K, and extending the setting from bandits to Markov decision processes to capture sequential decision making.


\newpage

\appendix

\textbf{Summary:}
The appendix is divided into three parts.

\begin{itemize}
\item Sections~\ref{app:easylemma-proof} to \ref{sec:subGaussian-bound-proof} contain the proofs of all the theorems and lemmas stated in the main body.

\item
Sections~\ref{app:gaussian-width}-\ref{app:bound-maxgaussian} present auxiliary tools used throughout the paper.

\item Sections~\ref{app:diameter}-\ref{sec:gibbs} include additional examples and a numerical study of Algorithm~\ref{alg:smallvertices}. Two settings are explored:
\begin{itemize}
    \item A low-dimensional ($n=3$) action space on the unit sphere, investigating the effect of cluster diameter on expected regret in line with Theorem~\ref{thm:geometric-bounds}.
    \item Nonstationary dependence over the action space via a nonstationary Gibbs kernel, where the kernel value depends not only on the difference $a - a'$ but also on the location of $a$ itself.
\end{itemize}

\end{itemize}

\section{Proof of Lemma~\ref{lem:measure-epsilon-net}}
\label{app:easylemma-proof}

\textbf{Statement:}
Given a partition \( \mathcal{R} \) of the full action space, and the importance measure \( q \) assigning a value to each cluster \(r\in\mathcal{R}\).
Let $\mathcal{A}$ be the output of Algorithm~\ref{alg:smallvertices} after $K$ samples. Then, with probability at least $1-\frac{1}{\epsilon} \exp(-K\epsilon)$, it holds that for any cluster $r\in\mathcal{R}$,
\[
r \cap  \mathcal{A} \neq \emptyset \quad \text{whenever} \quad q(r) > \epsilon.
\]
\begin{proof}
By definition, the algorithm draws $K$ i.i.d. samples from clusters in $\mathcal{R}$ according to the distribution $q$. Define a set of typical clusters
$R_{\epsilon}:=\{r\in\mathcal{R}: q(r)>\epsilon\}$.
Then, 
\begin{equation}
\Pr[r\cap\mathcal{A}\!=\!\emptyset]=(1-q(r))^K< (1-\epsilon)^K \leq \exp(-K \epsilon), \;\forall r\in R_{\epsilon},\label{equ:measure-epsilon-net-upperbound}
\end{equation}
where the equality is the probability of missing the cluster $r$ for $K$ times, the first inequality uses the definition of $R_{\epsilon}$. The last inequality uses $\log(1-\epsilon)^K=K\log(1-\epsilon)$ and the inequality $\log(1-\epsilon)\leq -\epsilon$ for $\epsilon<1$ \citep{topsoe12007some}.
Therefore, 
\begin{align*}
&\Pr[r \cap  \mathcal{A} \neq \emptyset \; \text{whenever} \; q(r) > \epsilon]\\
= & 1- \Pr[r \cap  \mathcal{A} \neq \emptyset \; \text{for some } r\in R_{\epsilon}]\\
\geq & 1-\sum_{r\in R_{\epsilon}} \Pr[r \cap  \mathcal{A} \neq \emptyset]\\
> & 1- \sum_{r\in R_{\epsilon}} \exp(-K \epsilon)\\
=& 1- |R_{\epsilon}|\exp(-K \epsilon)\geq 1- 1/\epsilon\cdot\exp(-K \epsilon),
\end{align*}
where the first inequality uses union bound, the second inequality uses \eqref{equ:measure-epsilon-net-upperbound}. The last inequality uses the fact that there could be at most $\lfloor 1/\epsilon \rfloor$ clusters in $R_{\epsilon}$, because $q$ is a probability measure and each $r\in\mathcal{R}_{\epsilon}$ has $q(r)>\epsilon$.
\end{proof}

\section{Proof of Theorem~\ref{thm:geometric-bounds}}

To bound the expected regret of reference subsets, we need a few more lemmas:
\begin{lemma}[Expectation integral identity]\label{lem:integral}
Given a non-negative random variables $X$. If $\Pr[X\geq u]\leq c\exp{\left(-\frac{u^2}{\epsilon^2}\right)}$ for any $u>0$, then $\E X\leq C\epsilon\sqrt{\log c}$,  where $\epsilon,c,C>0$ are positive constants.
\begin{proof}
\begin{align*}
\E  X 
=& \int_0^{\infty}\Pr[X\geq u]\; du \\
=& \int_0^{u_0}\Pr[X\geq u]\; du + \int_{u_0}^{\infty}\Pr[X\geq u]\; du\\
\leq & u_0 + \frac{1}{u_0} \int_{u_0}^{\infty}u\cdot\Pr[X\geq u]\; du\\
\leq & u_0 + \frac{c}{u_0} \int_{u_0}^{\infty}u\cdot \exp{\left(-\frac{u^2}{\epsilon^2}\right)}\; du\\
= & u_0 + \exp\left(-\frac{u_0^2}{\epsilon^2}\right)\cdot\frac{c\epsilon^2}{2u_0}\\
= & \epsilon\sqrt{\log c} + \frac{\epsilon}{2\sqrt{\log c}} \leq C \epsilon\sqrt{\log c},
\end{align*}
where the first equality uses integrated tail formula of expectation (cf. Lemma~1.6.1 of \citep{vershynin2018high}). The last equality set $u_0:=\epsilon\sqrt{\log c}$.
\end{proof}
\end{lemma}

\begin{lemma}[Borell-TIS inequality; Lemma~2.4.7 of \citep{talagrand2014upper}]\label{lem:Borell-TIS}
Given a set $\mathcal{S}$, and a zero-mean Gaussian process $\left(X_a\right)_{a\in\mathcal{S}}$. Let $\epsilon:=\sup_{a\in\mathcal{S}}\left(\E  X_a^2\right)^{\frac{1}{2}}$. Then for $u>0$, we have 
\begin{equation*}
\Pr\left[\left|\sup_{a\in\mathcal{S}} X_a - \E \sup_{a\in\mathcal{S}} X_a \right|\geq u\right]\leq 2\exp{\left(-\frac{u^2}{2\epsilon^2}\right)}.
\end{equation*}
\end{lemma}
It means that the size of the fluctuations of $\sup_{a\in\mathcal{S}} X_a $ is governed by the size of the individual random variables $X_a$. 

\paragraph{Statement of Theorem~\ref{thm:geometric-bounds}:}
Consider a partition \( \mathcal{R} := \{r_{\ell}\}_{\ell \leq m} \) of the full action space, with $\epsilon:=\max_{r\in\mathcal{R}}\diam(r)$, and an arbitrary reference subset $\mathcal{A}$. 
Then, for some constant \( C > 0 \),
\begin{equation*}
\E _{\theta}[\regret] \leq \max_{r\in\mathcal{R}}\E _{\theta}\left[\max_{a\in r} \mu_a - \mu_{a_{\ell}}\right] + C\epsilon\sqrt{\log m} \;=\;\max_{r\in\mathcal{R}}\E _{\theta}\left[\max_{a\in r} \mu_a \right] + C\epsilon\sqrt{\log m}.
\end{equation*}

\begin{proof}
Fix $\ell$, define a Gaussian process $\left\{Z_a\right\}_{a\in r_{\ell}}$ where $Z_a:=\mu_a-\mu_{a_{\ell}}$. Define a non-negative random variable $$Y_{\ell}:=\sup_{a\in r_{\ell}} Z_a=\sup_{a\in r_{\ell}} \mu_a-\mu_{a_{\ell}}.$$
Since $\E \mu_a=0$ for all $a\in\mathcal{A}_{\full}$, it holds that $\E Y_{\ell}=\E \max_{a\in r_{\ell}} \mu_a$. 

Further, by definition \( r_{\ell} \subset B(a_{\ell}, \epsilon) \), such that $\E  Z_a^2 = \E (\mu_a-\mu_{a_{\ell}})^2=\|a-a_{\ell}\|^2_2\leq \epsilon^2$.
Using Lemma~\ref{lem:Borell-TIS} on the process $\left\{Z_a\right\}_{a\in r_{\ell}}$, we have
\begin{equation*}
\Pr\left[\left|Y_{\ell} - \E Y_{\ell} \right|\geq u\right]\leq 2\exp{\left(-\frac{u^2}{2\epsilon^2}\right)}.
\end{equation*}
By union bound, we have
\begin{equation*}
\Pr\left[\max_{\ell\leq m}\left|Y_{\ell} - \E Y_{\ell} \right|\geq u\right]\leq  2m\exp{\left(-\frac{u^2}{2\epsilon^2}\right)}.
\end{equation*}
Using Lemma~\ref{lem:integral}, we have for some absolute constant $C>0$
\begin{equation}
\E \max_{\ell\leq m}\left|Y_{\ell} - \E Y_{\ell} \right| \leq C\epsilon\sqrt{\log m}.\label{equ:max-deviation}
\end{equation}

\noindent\textbf{Upper bound:}

When the cluster $r_{\ell}$ contains the optimal action, i.e., $a^*(\theta) \in r_{\ell}$, we have
\begin{equation}
\regret = \max_{a \in \mathcal{A}_{\full}} \mu_a - \max_{a' \in \mathcal{A}} \mu_{a'} = \max_{a \in r_{\ell}} \mu_a - \max_{a' \in \mathcal{A}} \mu_{a'} \leq \max_{a \in r_{\ell}} \mu_a - \mu_{a_{\ell}} = Y_{\ell} \leq \max_{\ell' \leq m} Y_{\ell'},\label{equ:regret-range}
\end{equation} 
where the first equality follows from the definition of regret in \eqref{equ:regret-define}, and the second equality follows from the assumption that $a^*(\theta) \in r_{\ell}$. The inequality holds because $a_{\ell} \in \mathcal{A}$, and hence $\max_{a \in \mathcal{A}} \mu_a \geq \mu_{a_{\ell}}$. The final equality follows from the definition of $Y_{\ell}$.
Therefore, 
\begin{align}
\E \regret\leq \E \max_{\ell\leq m}Y_{\ell}\leq \max_{\ell\leq m}\E Y_{\ell}+\E \max_{\ell\leq m}\left|Y_{\ell} - \E Y_{\ell} \right|\leq \max_{\ell\leq m}\E \max_{a \in r_{\ell}} \mu_a + C\epsilon\sqrt{\log m}, \label{equ:max-Yl}
\end{align}
where the first inequality uses \eqref{equ:regret-range}.
The second inequality uses $\max_{\ell\leq m}Y_{\ell}\leq \max_{\ell\leq m}\E Y_{\ell}+\max_{\ell\leq m}\left|Y_{\ell} - \E Y_{\ell} \right|$, since $Y_{\ell}\leq \E Y_{\ell}+\left|Y_{\ell} - \E Y_{\ell} \right|$.
The last inequality uses \eqref{equ:max-deviation} and the identity \( \E [Y_{\ell}] = \E [\max_{a \in r_{\ell}} \mu_a] \).
\end{proof}

\section{Proof of Theorem~\ref{thm:geometric-bounds-lower}}

The proof closely follows that of the upper bound, but introduces auxiliary variables $Y_{\ell,\ell'}$ indexed by pairs of clusters (allowing $\ell=\ell'$), instead of the single-index variables $Y_\ell$ used in the upper bound. This modification is necessary because, when $a^*(\theta) \in r_\ell$, the argument must compare against the best representative $a_{\ell'}$, i.e., the one attaining $\max_{\ell' \le m} \mu_{a_{\ell'}}$.

We define the outer radius induced by an action subset $\mathcal{S}$ as
\[
\Delta(\mathcal{S}):=\max_{a\in\mathcal{S}}\sup_{a'\in\mcAF}\|a-a'\|_2,
\]
which satisfies $\Delta(\mathcal{S})\le\diam(\mcAF)$.

\paragraph{Statement of Theorem~\ref{thm:geometric-bounds-lower}:}
Consider a partition \( \mathcal{R} := \{r_{\ell}\}_{\ell \leq m} \) of the full action space, with an arbitrary reference subset $\mathcal{A}^{\epsilon}$. 
Then, for some absolute constant \( C > 0 \),
\begin{align*}
\mathbb{E}_{\theta}[\regret]
\;\ge\;
\max_{\ell \le m}\E\left[ \max_{a\in r_{\ell}} \mu_a-\mu_{a_{\ell}} \right]
-
C\;\Delta(\mathcal{A}^{\epsilon})\sqrt{\log m}.
\end{align*}

\begin{proof}
\textbf{Step 1:}
Fix $\ell,\ell'$ (allowing $\ell=\ell'$), define a centered Gaussian process $\left\{Z_a\right\}_{a\in r_{\ell}\cup\{a_{\ell'}\}}$ where $Z_a:=\mu_a-\mu_{a_{\ell}}$. 
Define a non-negative random variable \[
Y_{\ell,\ell'}:=\sup_{a\in r_{\ell}\cup\{a_{\ell'}\}} Z_a=\sup_{a\in r_{\ell}\cup\{a_{\ell'}\}} \mu_a - \mu_{a_{\ell'}}.
\]

\textbf{Step 2:}
Set $\Delta:=\Delta(\mathcal{A}^{\epsilon})$.
Further, by definition of $\Delta$, we have $\E  Z_a^2 = \E (\mu_a-\mu_{a_{\ell}})^2=\|a-a_{\ell}\|^2_2\leq \Delta^2$.
Using Lemma~\ref{lem:Borell-TIS} on the process $\left\{Z_a\right\}_{a\in r_{\ell}\cup\{a_{\ell'}\}}$, we have
\begin{equation*}
\Pr\left[\left|Y_{\ell,\ell'} - \E Y_{\ell,\ell'} \right|\geq u\right]\leq 2\exp{\left(-\frac{u^2}{2\Delta^2}\right)}.
\end{equation*}
By union bound, we have
\begin{equation*}
\Pr\left[\max_{\ell,\ell'\le m}\left|Y_{\ell,\ell'} - \E Y_{\ell,\ell'} \right|\geq u\right]\leq 2m^2\exp{\left(-\frac{u^2}{2\Delta^2}\right)}.
\end{equation*}
Using Lemma~\ref{lem:integral}, we have for some absolute constant $C>0$
\begin{equation}
\E \max_{\ell,\ell'\le m}\left|Y_{\ell,\ell'} - \E Y_{\ell,\ell'} \right| \leq C\Delta\sqrt{2\log m}.\label{equ:max-deviation-1}
\end{equation}

\textbf{Step 3:}
Now, we connect these auxiliary variables $Y_{\ell,\ell'}$ to regret definition.
Fix $\ell\le m$. Let the cluster contains the optimal action, i.e., $a^*(\theta)\in r_{\ell}$, we have
\begin{align*}
\regret 
&= \max_{a \in \mathcal{A}_{\full}} \mu_a - \max_{a' \in \mathcal{A}} \mu_{a'} \\
&=\max_{a \in r_{\ell}} \mu_a - \max_{\ell'\le m} \mu_{a_{\ell'}}\\
&=\min_{\ell'\le m} \left\{\max_{a \in r_{\ell}} \mu_a- \mu_{a_{\ell'}}\right\}\\
&=\min_{\ell'\le m} \left\{\max_{a \in r_{\ell}\cup\{a_{\ell'}\}} \mu_a- \mu_{a_{\ell'}}\right\}\\
&= \min_{\ell'\le m} Y_{\ell,\ell'}\\
&= \min_{\ell'\le m} \max_{\ell''\le m} Y_{\ell'',\ell'}
\end{align*}
The second equality uses $a^*(\theta)\in r_{\ell}$ and $\mcA=\{a_{\ell}\}_{\ell\le m}$ is the reference subset.
The fourth equality uses under the event of $a^*(\theta)\in r_{\ell}$, we have
\[
\max_{a\in \mcAF} \mu_a=\max_{a\in r_{\ell}} \mu_a \ge \mu_{a_{\ell'}}
\]
such that $\max_{a \in r_{\ell}\cup\{a_{\ell'}\}} \mu_a= \max_{a \in r_{\ell}} \mu_a$.
The fifth equality uses the definition of $Y_{\ell,\ell'}$.
The last uses that under the event of $a^*(\theta)\in r_{\ell}$, we have $\max_{a\in r_{\ell}\cup\{a_{\ell'}\}}\mu_a \ge \max_{a\in r_{\ell''}\cup\{a_{\ell'}\}}\mu_a$ for any $\ell''\le m$, such that
\[
Y_{\ell,\ell'}\ge Y_{\ell'',\ell'} \quad\forall \ell''\le m.
\]
Therefore, since $\ell$ is included in $1,\dots,m$, we reach the fact that
\[
Y_{\ell,\ell'}=\max_{\ell''\le m} Y_{\ell'',\ell'}. 
\]
Note that this term $\min_{\ell'\le m} \max_{\ell''\le m} Y_{\ell'',\ell'}$ is already irrelevant to $\ell$. We conclude that for any $\theta$,
\[
\regret \ge \min_{\ell'\le m} \max_{\ell\le m} Y_{\ell,\ell'}. 
\]

\textbf{Step 4:}
Using $Y_{\ell, \ell'} = \E Y_{\ell, \ell'} + (Y_{\ell, \ell'}- \E Y_{\ell, \ell'})$, we have
\begin{equation}
\begin{aligned}
\min_{\ell' \le m}\max_{\ell \le m} Y_{\ell,\ell'}
&=
\min_{\ell' \le m}\max_{\ell \le m}
\Bigl(
\mathbb{E}Y_{\ell,\ell'} + (Y_{\ell,\ell'}-\mathbb{E}Y_{\ell,\ell'})
\Bigr) \\
&\ge
\min_{\ell' \le m}\max_{\ell \le m}
\Bigl(
\mathbb{E}Y_{\ell,\ell'} - |Y_{\ell,\ell'}-\mathbb{E}Y_{\ell,\ell'}|
\Bigr) \\
&\ge
\min_{\ell' \le m}\max_{\ell \le m}
\Bigl(
\mathbb{E}Y_{\ell,\ell'} -
\max_{\ell'' \le m}\max_{\ell''' \le m}
|Y_{\ell'',\ell'''}-\mathbb{E}Y_{\ell'',\ell'''}|
\Bigr) \\
&=
\min_{\ell' \le m}\max_{\ell \le m}\mathbb{E}Y_{\ell,\ell'}
-
\max_{\ell'' \le m}\max_{\ell''' \le m}
|Y_{\ell'',\ell'''}-\mathbb{E}Y_{\ell'',\ell'''}|.
\end{aligned}
\label{equ:min-max-decomp}
\end{equation}
The first uses the decomposition
\(Y_{\ell,\ell'}=\mathbb{E}Y_{\ell,\ell'}+(Y_{\ell,\ell'}-\mathbb{E}Y_{\ell,\ell'})\).
The second follows from the inequality \(x \ge x - |x|\), applied point-wise.
For the third line, we use that for all \((\ell,\ell')\),
\[
|Y_{\ell,\ell'}-\mathbb{E}Y_{\ell,\ell'}|
\le
\max_{\ell'' \le m}\max_{\ell''' \le m}
|Y_{\ell'',\ell'''}-\mathbb{E}Y_{\ell'',\ell'''}|,
\]
so each absolute value can be uniformly upper bounded. In the last step, the
term
\[
\max_{\ell'' \le m}\max_{\ell''' \le m}
|Y_{\ell'',\ell'''}-\mathbb{E}Y_{\ell'',\ell'''}|
\]
does not depend on \(\ell,\ell'\), so it can be taken outside the inner maximum.

\textbf{Step 5: Min-max swap}
For every fixed pair \((\ell,\ell')\), one has
\[
\min_{j \le m}\mathbb{E}Y_{\ell,j}
\;\le\;
\mathbb{E}Y_{\ell,\ell'}
\;\le\;
\max_{i \le m}\mathbb{E}Y_{i,\ell'}.
\]
Since this holds for every \(\ell,\ell'\), it follows that for every \(\ell,\ell'\),
\[
\min_{j \le m}\mathbb{E}Y_{\ell,j}
\;\le\;
\max_{i \le m}\mathbb{E}Y_{i,\ell'}.
\]
Now taking \(\max_{\ell \le m}\) on the left-hand side gives
\[
\max_{\ell \le m}\min_{j \le m}\mathbb{E}Y_{\ell,j}
\;\le\;
\max_{i \le m}\mathbb{E}Y_{i,\ell'}
\qquad\text{for every }\ell' \le m.
\]
Finally, taking \(\min_{\ell' \le m}\) on the right-hand side yields
\[
\max_{\ell \le m}\min_{\ell' \le m}\mathbb{E}Y_{\ell,\ell'}
\;\le\;
\min_{\ell' \le m}\max_{\ell \le m}\mathbb{E}Y_{\ell,\ell'}.
\]
Hence, we have
\begin{equation}
\min_{\ell' \le m}\max_{\ell \le m}\mathbb{E}Y_{\ell,\ell'}
\;\ge\;
\max_{\ell \le m}\min_{\ell' \le m}\mathbb{E}Y_{\ell,\ell'}.
\label{equ:min-max-reverse}
\end{equation}

\textbf{Step 6:}
Recall the definition of $Y_{\ell,\ell'}$, fix $\ell$
\begin{equation}
\min_{\ell' \le m}\mathbb{E}Y_{\ell,\ell'} 
= \min_{\ell' \le m}\E\left[ \max_{a\in r_{\ell}\cup\{a_{\ell'}\}} \mu_a - \mu_{a_{\ell'}}\right]=\min_{\ell' \le m}\E\left[ \max_{a\in r_{\ell}\cup\{a_{\ell'}\}} \mu_a \right]=\E\left[ \max_{a\in r_{\ell}} \mu_a \right]
=\E\left[ \max_{a\in r_{\ell}} \mu_a-\mu_{a_{\ell}} \right],
\label{equ:min-EYll}
\end{equation}
where the first equality use the definition of $Y_{\ell,\ell'}$. The second uses $\E \mu_{a_{\ell'}}=0$.
The third uses 
\(
\max_{a\in r_{\ell}\cup\{a_{\ell'}\}} \mu_a  \ge \max_{a\in r_{\ell}} \mu_a, \forall \ell'\le m.
\)
The last uses again $\E \mu_{a_{\ell}}=0$.

\textbf{Step 7: Wrap up.}
\begin{align*}
\E[\regret]
&\ge \min_{\ell' \le m}\max_{\ell \le m}\mathbb{E}Y_{\ell,\ell'}
-
\E\max_{\ell'' \le m}\max_{\ell''' \le m}
|Y_{\ell'',\ell'''}-\mathbb{E}Y_{\ell'',\ell'''}|  \\
&\ge \max_{\ell \le m}\min_{\ell' \le m}\mathbb{E}Y_{\ell,\ell'}
-
\E\max_{\ell'' \le m}\max_{\ell''' \le m}
|Y_{\ell'',\ell'''}-\mathbb{E}Y_{\ell'',\ell'''}|\\
&\ge \max_{\ell \le m}\E\left[ \max_{a\in r_{\ell}} \mu_a-\mu_{a_{\ell}} \right]
-C\Delta\sqrt{2\log m}.
\end{align*}
The first use Equation~\eqref{equ:min-max-decomp}.
The second uses Equation~\eqref{equ:min-max-reverse}.
The last uses Equations~\eqref{equ:min-EYll} and~\eqref{equ:max-deviation-1}.  Let $C'=C\sqrt{2}$ completes the proof.
\end{proof}

\section{Proof of Theorem~\ref{thm:alg-bound-upper}}

An important consequence of assuming a centered Gaussian process, and thus
$\E _{\theta}\left[\mu_a\right]=0$ for all $a\in\mathcal{A}_{\full}$ is the following property:
\begin{lemma}[Transformation invariance]\label{lem:invariant-transform}
Given any vector $x\in\mathbb{R}^n$, let $S+x:=\{s+x:s\in\mathcal{S}\}$. If $\E [\theta]=0$, then \(\E \left[\max_{a\in\mathcal{S}+x}\langle a, \theta\rangle\right]=\E \left[\max_{a'\in\mathcal{S}}\langle a', \theta\rangle\right]\).
\begin{proof}
\begin{align*}
\E \left[\max_{a\in\mathcal{S}+x}\langle a, \theta\rangle\right]
=&\E \left[\max_{a'\in\mathcal{S}}\langle a'+x, \theta\rangle\right]\\
=&\E \left[\max_{a'\in\mathcal{S}}\langle a', \theta\rangle\right] +\E \left[\langle x, \theta\rangle\right]\\
=&\E \left[\max_{a'\in\mathcal{S}}\langle a', \theta\rangle\right],
\end{align*}
where the last equality uses $\E \left[\langle x, \theta\rangle\right]=\left\langle x,\E [\theta]\right\rangle=0$.
\end{proof}
\end{lemma}

The key tool for decomposing the regret of Algorithm~\ref{alg:smallvertices} into that from a reference subset and that from missing clusters is the following Lemma:

\begin{lemma}\label{lem:bound-probability}
Consider a partition \( \mathcal{R} \) of the full action space.  
Let \( \mathcal{A} \) be the output of Algorithm~\ref{alg:smallvertices}.  
Then, the event that the optimal action falls in a cluster \( r \), i.e., \( \{a^*(\theta) \in r\} \), is independent of whether the subset \( \mathcal{A} \) intersects with the cluster.
It holds that
\begin{align*}
\Pr[a^*(\theta)\in r,\, r\cap\mathcal{A}\!=\!\emptyset] 
=&  q(r)(1-q(r))^K,\\
\Pr[a^*(\theta)\in r, r\cap\mathcal{A}\!\neq\!\emptyset] 
\leq&  q(r).
\end{align*}
\begin{proof}
Since $\mathcal{A}$ is the output of Algorithm~\ref{alg:smallvertices}, the event $\{a^*(\theta)\in r\}$ is independent from $\{r\cap\mathcal{A}\!\neq\!\emptyset\}$ or $\{r\cap\mathcal{A}\!=\!\emptyset\}$.
We have
\begin{align*}
\Pr[a^*(\theta)\in r, r\cap\mathcal{A}\!=\!\emptyset]
=&\Pr[a^*(\theta)\in r]\Pr[r\cap\mathcal{A}\!=\!\emptyset]\\
=&q(r)\Pr[r\cap\mathcal{A}\!=\!\emptyset]\\
=&q(r) (1-q(r))^K,
\end{align*}
where the first equality uses independence between $\{a^*(\theta)\in r\}$ and $\{r\cap\mathcal{A}\!\neq\!\emptyset\}$, the second equality uses the definition of measure $q$, the third equality use the probability of missing cluster $r$ in $K$ i.i.d. samplings. Similarly, 
\begin{align*}
\Pr[a^*(\theta)\in r, r\cap\mathcal{A}\!\neq\!\emptyset]=q(r)\cdot \Pr[r\cap\mathcal{A}\!\neq\!\emptyset].
\end{align*}
Using $\Pr[r\cap\mathcal{A}\!\neq\!\emptyset]\leq 1$, we complete the proof.
\end{proof}
\end{lemma}

\paragraph{Statement of Theorem~\ref{thm:alg-bound-upper}:}
Let $\mathcal{A}$ be output of Algorithm~\ref{alg:smallvertices}. 
Consider a partition \( \mathcal{R} := \{r_{\ell}\}_{\ell \leq m} \) of the full action space, with $\epsilon:=\max_{r\in\mathcal{R}}\diam(r)$.
Then, for the same constant \( C > 0 \) in Theorem~\ref{thm:geometric-bounds},
\begin{equation*}
\begin{split}
\E _{\theta,\mathcal{A}}[\textnormal{Regret}] 
\leq & 
\max_{r\in\mathcal{R}}\E _{\theta}\left[\max_{a \in r} \mu_a\right] + C\epsilon\sqrt{\log m} \\
&+ \left(\E _{q}\left[(1-q(r))^{2K}\right]\cdot \E _{\theta}\left[\max_{a \in \mathcal{A}_{\full}} \mu^2_a\right]\right)^{1/2}.
\end{split}
\end{equation*}

\begin{proof}
If \( r_{\ell} \cap \mathcal{A} \neq \emptyset \), define \( a_{\ell} \in r_{\ell} \cap \mathcal{A} \). 
If $r_{\ell}\cap\mathcal{A}\!=\!\emptyset$, choose an arbitrary point $a_{\ell}\in r_{\ell}$ as the representative.
The set $\mathcal{A}':=\{a_{\ell}\}_{\ell\leq m}$ forms a reference subset of Definition~\ref{def:ref-subset}.
The cluster \( r_{\ell} \) is contained in a closed Euclidean ball of radius \( \epsilon \) centered at \( a_{\ell} \), i.e., \( r_{\ell} \subset B(a_{\ell}, \epsilon) \). 
Define a Gaussian process $\left\{Z_a\right\}_{a\in r_{\ell}}$ where $Z_a:=\mu_a-\mu_{a_{\ell}}$. Define a random variable $$Y_{\ell}:=\sup_{a\in r_{\ell}} Z_a=\sup_{a\in r_{\ell}} \mu_a-\mu_{a_{\ell}}.$$
Since $\E \mu_a=0$ for all $a\in\mathcal{A}_{\full}$, we have $\E Y_{\ell}=\E \sup_{a\in r_{\ell}} \mu_a$.

Consider the case that $a^*(\theta)\in r_{\ell}$. We have
\begin{equation}
\begin{split}
&\E \left[\regret \Big| r_{\ell}\cap\mathcal{A}\!\neq\!\emptyset,a^*(\theta)\in r_{\ell} \right]\\
\leq & \E \left[ Y_{\ell} \Big| r_{\ell}\cap\mathcal{A}\!\neq\!\emptyset,a^*(\theta)\in r_{\ell} \right]\\
=& \E \left[ Y_{\ell} \Big|a^*(\theta)\in r_{\ell} \right],\\
\leq & \E \left[\max_{\ell \leq m}Y_{\ell}\Big|a^*(\theta)\in r_{\ell}\right],
\end{split}
\label{equ:alg-bound-1}
\end{equation}
where the first inequality uses $\regret\leq Y_{\ell}$ if $r_{\ell}\cap\mathcal{A}\!\neq\!\emptyset$, the equality uses that $Y_{\ell}$ is independent of $r_{\ell}\cap\mathcal{A}\!\neq\!\emptyset$. 
On the other hand, we assume that $0\in\mathcal{A}$ because even if it doesn't hold, we can always find a vector $x\in\mathbb{R}^n$ such that $0\in\mathcal{A}+x$, without changing the value of $\E [\max_{a\in\mathcal{A}}\mu_a]$ (c.f. Lemma~\ref{lem:invariant-transform}).
This is useful because $\max_{a\in\mathcal{A}}\mu_a(\theta)\ge \langle 0,\theta\rangle=0$.
Therefore,  
\begin{align}
\begin{split}
&\E \left[\regret \Big| r_{\ell}\cap\mathcal{A}\!=\!\emptyset,a^*(\theta)\in r_{\ell} \right]\\
\leq &\E \left[\max_{a\in\mathcal{A}_{\full}} \mu_a \Big| a^*(\theta)\in r_{\ell}\right],\label{equ:alg-bound-2}
\end{split}
\end{align}
where the inequality uses $\regret\leq \max_{a\in\mathcal{A}_{\full}} \mu_a$, as a consequence of $0\in\conv(\mathcal{A})$, and that $\max_{a\in\mathcal{A}_{\full}} \mu_a $ is independent of $r_{\ell}\cap\mathcal{A}\!=\!\emptyset$.
Further,
\begin{equation}
\begin{aligned}
\E [\regret] = & \sum_{r\in\mathcal{R}} \Pr[r \cap\mathcal{A}\neq\emptyset,a^*(\theta)\in r ]\cdot\E \left[\regret \Big| r \cap\mathcal{A}\neq\emptyset,a^*(\theta)\in r  \right]\\
& + \sum_{r\in\mathcal{R}} \Pr[r \cap\mathcal{A}=\emptyset,a^*(\theta)\in r ]\cdot\E \left[\regret \Big| r \cap\mathcal{A}=\emptyset,a^*(\theta)\in r  \right]\\
\leq & \sum_{\ell\leq m} q(r_{\ell})\cdot \left(\E \left[\max_{\ell\leq m} Y_{\ell} \Big|a^*(\theta)\in r_{\ell} \right] +  (1-q(r_{\ell}))^K\cdot \E \left[\max_{a\in\mathcal{A}_{\full}} \mu_a\Big|a^*(\theta)\in r_{\ell}\right] \right) \\
= & \E \left[\max_{\ell\leq m} Y_{\ell}\right] + \sum_{r\in\mathcal{R}} q(r )\cdot (1-q(r))^K\cdot \E \left[\max_{a\in\mathcal{A}_{\full}} \mu_a\Big|a^*(\theta)\in r \right] \\
\leq & \E \left[\max_{\ell\leq m} Y_{\ell}\right] + \left(\E _{q}\left[(1-q(r))^{2K}\right]\cdot \E _{\theta}\left[\max_{a \in \mathcal{A}_{\full}} \mu^2_a\right]\right)^{1/2},
\end{aligned}
\label{equ:EY+extra}
\end{equation}
where the first equality uses tower rule.
The first inequality uses Equations~(\ref{equ:alg-bound-1}-\ref{equ:alg-bound-2}), and Lemma~\ref{lem:bound-probability}.
The last equality uses tower rule again.
The last inequality uses Cauchy-Schwarz inequality, which states that for any two random variables $X$ and $Y$, we have $|\E [XY]|\leq\sqrt{\E [X^2]\E [Y^2]}$.

By applying \eqref{equ:max-Yl} in the proof of Theorem~\ref{thm:geometric-bounds}, we complete the proof.
\end{proof}

\section{Proof of Theorem~\ref{thm:alg-bound-upper-worst}}

To analyze the worst-case behavior of Theorem~\ref{thm:alg-bound-upper}, we first examine the term $\E _q\left[(1 - q(r))^K\right]$ in the upper bound, summarized in the following Lemma:

\begin{lemma}\label{lem:maximum-(1-q)^K}
Let $q$ denote a discrete probability distribution over a finite support $\mathcal{R}$. Define
\begin{equation}
\begin{split}
M := \max_{q} &\;\E _q\left[(1 - q(r))^K\right] \\
\textnormal{s.t.}& \sum_{r \in \mathcal{R}} q(r) = 1,\; q(r) \in [0,1) \;\;\forall r \in \mathcal{R}.     
\end{split}
\label{equ:opt-Eq(1-q)}
\end{equation}
When $m \geq K + 1$, the maximum is
$M = \left(1 - \frac{1}{m}\right)^K$,
and it is attained when $q$ is uniform.
When $m < K + 1$, the maximum is upper bounded by $M\leq \frac{m}{K+1} \left(\frac{K}{K+1}\right)^K$, and there exists a feasible distribution $q'$ such that
$\E _{q'}\left[(1 - q'(r))^K\right] \geq \frac{m - 1}{K+1} \left(\frac{K}{K+1}\right)^K$.
\begin{proof}
Let $\mathcal{R} = \{r_1, \dots, r_m\}$ be the finite support of the measure $q$, where $m := m$. Let $q_i := q(r_i)$ denote the probability mass at each support point.
Define the function $f(q_{\ell}) := q_{\ell} \cdot (1 - q_{\ell})^K$.

\textbf{Case I $m< K+1$: }
The first derivative of $f$ is
$$
f'(q_{\ell}) = (1 - q_{\ell})^{K - 1} \big(1 - (K + 1) q_{\ell}\big),
$$
which is positive on the interval $\left[0, \frac{1}{K+1}\right)$ and negative on the interval $\left(\frac{1}{K+1}, 1\right)$. Therefore, $f(q_{\ell})$ attains its maximum over $[0,1)$ at
$$
q^*_{\ell} = \frac{1}{K+1},
$$
with the corresponding maximum value
$$
 f(q^*_{\ell}) \leq \frac{1}{K+1} \left(\frac{K}{K+1}\right)^K.
$$
Since there are $m$ support points, this yields the upper bound
$
M\leq\sum_{\ell=1}^m f(q^*_{\ell}) \leq \frac{m}{K+1} \left(\frac{K}{K+1}\right)^K
$.
Also, since $\frac{1}{K+1}<\frac{1}{m}$, the solution $q_1=\dots=q_{m-1}=\frac{1}{K+1}$ and $q_{m}=\frac{K-m+2}{K+1}$ is feasible. Thus,
\begin{align*}
M\geq & \sum_{\ell=1}^{m-1} f\left(\frac{1}{K+1}\right)+f\left(\frac{K-m+2}{K+1}\right)\\
\geq &\frac{m-1}{K+1} \left(\frac{K}{K+1}\right)^K,
\end{align*}
where the right inequality uses that $f(q_{\ell})\geq 0$ for $q_{\ell}\in[0,1]$.

\textbf{Case II $m\geq K+1$: }
Consider the relaxed maximization problem of \eqref{equ:opt-Eq(1-q)}:
\begin{equation}
\max_{q_1,\dots,q_m \in [0,1)} \sum_{\ell=1}^m f(q_{\ell})\quad \text{subject to } \sum_{\ell=1}^m q_{\ell} \leq 1. \label{equ:relaxed-opt}
\end{equation}
Let $\lambda \geq 0$ be the Lagrange multiplier associated with the constraint. Define the Lagrangian:
$$
\mathcal{L}(q_1,\dots,q_m,\lambda) = \sum_{\ell=1}^m q_{\ell}(1 - q_{\ell})^K - \lambda\left(\sum_{\ell=1}^m q_{\ell} - 1\right).
$$
For each $\ell = 1, \dots, m$, compute the partial derivative of $\mathcal{L}$ with respect to $q_\ell$:
$$
\frac{\partial}{\partial q_{\ell}} \left[ q_{\ell}(1 - q_{\ell})^K \right] = (1 - q_{\ell})^K - K q_{\ell} (1 - q_{\ell})^{K - 1}.
$$
Setting this derivative equal to zero yields the stationary condition:
$$
(1 - q_{\ell})^{K - 1} (1 - (K + 1) q_{\ell}) = \lambda, \quad \text{with } \lambda \geq 0.
$$
To find critical points of \eqref{equ:relaxed-opt}, we solve the system:
$$
(1 - q_{\ell})^{K - 1} (1 - (K + 1) q_{\ell}) = \lambda \geq 0 \quad \forall \ell \leq m, \quad \text{and} \quad \sum_{\ell=1}^m q_{\ell} \leq 1.
$$
Define the function $g(q_{\ell}) := (1 - q_{\ell})^{K - 1} (1 - (K + 1) q_{\ell})$.
For $g(q_{\ell}) \geq 0$, it must hold that $1 - (K + 1) q_{\ell} \geq 0$, i.e., $q_{\ell} \leq \frac{1}{K + 1}$.
Therefore, any feasible solution to this system must satisfy $q_{\ell} \in \left[0,\frac{1}{K+1}\right]$ for all $\ell \leq m$.

Then, over the interval $q_{\ell} \in \left[0,\frac{1}{K+1}\right]$, both factors $(1 - q_{\ell})^{K - 1}$ and $1 - (K + 1) q_{\ell}$ are positive and decreasing. Hence, $g(q_{\ell})$ is positive and strictly decreasing, so the equation $g(q_{\ell}) = \lambda \geq 0$ has at most one solution.
Therefore, all $q_{\ell}$'s must be equal at a critical point. Let $q_{\ell} = c$ for all $\ell \leq m$. Due to the assumption of $m \geq K + 1$, we have $\frac{1}{m} \leq \frac{1}{K + 1}$, so any choice of $c\in\left[0,\frac{1}{m}\right]$ satisfies the constraint $\sum_{\ell=1}^m q_\ell \leq 1$ and is feasible for the system.

Therefore, $q_\ell = c$ for all $\ell \leq m$ is a feasible critical point of \eqref{equ:relaxed-opt}. The corresponding objective value is:
$$
\sum_{\ell=1}^m f(c)=m \cdot c \cdot (1 - c)^K,
$$
which is increasing in $c$ over the interval $\left[0,\frac{1}{m}\right]$. Hence, the maximum is attained at $c = \frac{1}{m}$, and the optimal value is:
$$
\sum_{\ell=1}^m f\left(\frac{1}{m}\right)=\left(1 - \frac{1}{m}\right)^K = \left(1 - \frac{1}{m}\right)^K,
$$
achieved when $q_{\ell} = \frac{1}{m}$ for all $\ell$.

To confirm that this critical point is indeed a maximum, observe that the Hessian of the objective function $f(q_{\ell})$ is diagonal (since the function is separable), and the diagonal entries are:
$$
\frac{\partial^2}{\partial q_{\ell}^2} \left[ q_{\ell}(1 - q_{\ell})^K \right] = -(1 - q_{\ell})^{K - 2} \left( 2K - K(K + 1) q_{\ell} \right),
$$
which is negative for $q_{\ell} \leq \frac{1}{K + 1}$, because then $2 - (K + 1)q_{\ell} > 0$. Hence, the Hessian is negative definite, and the critical point is a local (and thus global) maximum.

Finally, note that \eqref{equ:relaxed-opt} is a relaxation of \eqref{equ:opt-Eq(1-q)}. While the optimum of the original problem is upper bounded by that of the relaxed problem, the optimal solution to the relaxed problem also lies within the feasible region of the original problem. Therefore, the maximum of \eqref{equ:opt-Eq(1-q)} is also attained when $q$ is uniform, with the maximum value being $\left(1 - \frac{1}{m} \right)^K$.
\end{proof}
\end{lemma}

\begin{assumption}\label{ass:compact}
The support of measure $q$ is compact.
\end{assumption}
The compactness assumption ensures that the term $\E _{\theta}\left[\max_{a \in \mathcal{A}_{\full}} \mu_a \right]$ is finite and guarantees the attainment of a unique optimal action. Without loss of generality, we assume that $\mathcal{A}_{\full}$ is the support of the measure $q$.

\paragraph{Statement of Theorem~\ref{thm:alg-bound-upper-worst}}
Under Assumption~\ref{ass:compact}, there exists a point \( a_0 \) and a constant \( M > 0 \) such that \( \mathcal{A}_{\full} \subset B(a_0, M) \), a closed Euclidean ball of radius \( M \) centered at $a_0$. 
Let the action space have dimension \( n \in \mathbb{N} \), and fix a constant \( 0<\epsilon < M \).
Let \( \mathcal{A} \) be the output of Algorithm~\ref{alg:smallvertices}. For the same constant \( C> 0 \) in Theorem~\ref{thm:geometric-bounds}, we have:
\[
\E _{\theta,\mathcal{A}}[\regret] \leq 2\epsilon \min\{\sqrt{n},\sqrt{2\log |\mathcal{A}_{\full}|}+1\} + C\epsilon\sqrt{\log N(\mathcal{A}_{\full},\epsilon)}, \quad \text{where } K \geq \frac{1}{2}\left( \frac{M^2 N(\mathcal{A}_{\full},\epsilon)}{\epsilon^2 e} - 1 \right).
\]
As $\epsilon \to 0^+$, we have $\E _{\theta,\mathcal{A}}[\regret] \to 0$ as $K \to \infty$.
\begin{proof}
Let \( \{a_1, \dots, a_m\} \subseteq \mathcal{A}_{\full} \) be a minimal geometric \( \epsilon \)-net under the Euclidean norm, so that \( m = N(\mathcal{A}_{\full}, \epsilon) \). Define \( \pi(a) \) as the closest point in the $\epsilon$-net to \( a \), and let the partition \( \mathcal{R} = \{r_1, \dots, r_m\} \) be given by
\[
r_{\ell} = \left\{ a \in \mathcal{A}_{\full} : \pi(a) = a_{\ell} \right\}.
\]

\textbf{Step~1 bounding by dimension $n$:}

By Lemma~\ref{lem:invariant-transform}, we can shift the action space to be centered at the origin. So, without loss of generality, we assume that \( \mathcal{A}_{\full} \subset M\cdot B_2^n \), the scaled unit Euclidean ball in \( \mathbb{R}^n \). Then,
\begin{equation}
\E _{\theta}\left[\max_{a \in \mathcal{A}_{\full}} \mu_a^2\right] \leq M^2\cdot \E \|\theta\|_2^2 = M^2 n, \label{equ:wbound-1}
\end{equation}
where the inequality follows from \( \mu_a = \langle \theta, a \rangle \leq M\|\theta\|_2 \) for all \( a \in M B_2^n \), and the equality uses that \( \theta \sim \mathcal{N}(0, I) \), so each coordinate has variance 1 and \( \|\theta\|_2^2 = \sum_{i=1}^n \theta_i^2 \) has expectation \( n \).

Then,
\begin{equation}
\max_{\ell \leq m} \E _{\theta}\left[\max_{a \in r_\ell} \mu_a\right] 
\leq \E _{\theta}\left[\max_{a \in B(a_\ell, \epsilon)} \mu_a\right] 
= \E _{\theta}\left[\max_{a \in \epsilon B_2^n} \mu_a\right] 
\leq \epsilon \sqrt{n},\label{equ:wbound-2}
\end{equation}
where the first inequality follows from \( r_\ell \subset B(a_\ell, \epsilon) \) by construction; the equality uses Lemma~\ref{lem:invariant-transform} to shift; and the final bound uses Property~3 from Appendix~\ref{app:gaussian-width}.

\textbf{Step~2 bounding by cardinality $|\mathcal{A}_{\full}|$:}

By the definition of variance:
\begin{equation}
\begin{split}
\E _{\theta}\left[\max_{a \in \mathcal{A}_{\full}} \mu_a^2\right] 
=& \left(\E _{\theta}\left[\max_{a \in \mathcal{A}_{\full}} \mu_a\right]\right)^2 + \text{Var}\left(\max_{a \in \mathcal{A}_{\full}} \mu_a\right)\\
\leq & \left(M \sqrt{2\log |\mathcal{A}_{\full}|}\right)^2 + 4M^2\\
\le & M^2 \left(\sqrt{2\log |\mathcal{A}_{\full}|}+2\right)^2,
\end{split}
\end{equation}
where the inequality uses Appendix~\ref{app:bound-maxgaussian} to bound the square of Gaussian width, and Lemma~\ref{lem:Borell-TIS} to bound the variance.
The last equality uses the definition of $M$ and simple rearrangement.

Also,
\begin{equation}
\max_{\ell \leq m} \E _{\theta}\left[\max_{a \in r_\ell} \mu_a\right] 
\leq \max_{\ell \leq m} \diam(r_{\ell})\sqrt{2\log |r_{\ell}|}
\leq \epsilon \sqrt{2\log |\mathcal{A}_{\full}|},\label{equ:wbound-3}
\end{equation}
where the left inequality uses Appendix~\ref{app:bound-maxgaussian}, the right inequality uses the definition of $\eps$.

Then, 
\begin{align*}
\E _{\theta,\mathcal{A}}[\textnormal{Regret}] \leq &\max_{r\in\mathcal{R}}\E _{\theta}\left[\max_{a \in r} \mu_a\right] + C\epsilon\sqrt{\log m}  + \left(\E _{q}\left[(1-q(r))^{2K}\right]\cdot \E _{\theta}\left[\max_{a \in \mathcal{A}_{\full}} \mu^2_a\right]\right)^{1/2},\\
\leq & \epsilon\min\{\sqrt{n},\sqrt{2\log|\mathcal{A}_{\full}|}\} + C\epsilon\sqrt{\log m} \\
&+ \sqrt{\frac{m}{2K+1}}\left(\frac{2K}{2K+1}\right)^{K} M\;\min\{\sqrt{n},\sqrt{2\log |\mathcal{A}_{\full}|}+2\}\\
\leq & \epsilon\min\{\sqrt{n},\sqrt{2\log|\mathcal{A}_{\full}|}\} + C\epsilon\sqrt{\log m} + \epsilon \min\{\sqrt{n},\sqrt{2\log |\mathcal{A}_{\full}|}+2\}\\ 
\leq & 2\epsilon \min\{\sqrt{n},\sqrt{2\log |\mathcal{A}_{\full}|}+1\} + C\epsilon\sqrt{\log N(\mathcal{A}_{\full},\epsilon)},
\end{align*}
where the first inequality follows from the algorithm-dependent upper bound in Theorem~\ref{thm:alg-bound-upper}.  
The second inequality follows from Equations~\eqref{equ:wbound-1}-\eqref{equ:wbound-3}, and the term $\E _{q}\left[(1-q(r))^{2K}\right]$ is bounded by $\frac{m}{2K+1}\left(\frac{2K}{2K+1}\right)^{2K}$ (see Lemma~\ref{lem:maximum-(1-q)^K}).
The third inequality comes from
\begin{align*}
\log \left(\sqrt{\frac{m}{2K+1}}\left(\frac{2K}{2K+1}\right)^{K}M\right) = &\tfrac{1}{2} \log\left(\tfrac{m}{2K+1}\right) + K \log\left(\tfrac{2K}{2K+1}\right) + \log M\\
\leq &\tfrac{1}{2} \log m - \tfrac{1}{2} \log(2K+1) - \tfrac{1}{2} + \log M\\
\leq & \tfrac{1}{2} \log m - \tfrac{1}{2} \log\left( \frac{m M^2}{\epsilon^2 e} \right) - \tfrac{1}{2} + \log M\leq \log \epsilon,
\end{align*}
where the first inequality uses $\tfrac{1}{2} \log\left(\tfrac{m}{2K+1}\right) = \tfrac{1}{2}(\log m - \log(2K+1))$. Also, $\log\left(\tfrac{2K}{2K+1}\right) = \log\left(1 - \tfrac{1}{2K+1} \right) \leq -\tfrac{1}{2K+1}$ due to the inequality \( \log(1 - X) \leq -X \) for \( X < 1 \), and for a large $K$, we approximate $-\tfrac{K}{2K+1}\approx -\frac{1}{2}$.
The last inequality uses the assumption $K \geq \frac{1}{2}\left( \frac{m M^2}{\epsilon^2 e} - 1 \right)$. 

Finally, since \( \mathcal{A}_{\full} \subset M B_2^n \), the covering number satisfies \( N(\mathcal{A}_{\full}, \epsilon) \leq C' (M/\epsilon)^n \) for some constant \( C' > 0 \); see Proposition 4.2.12 of \citep{vershynin2018high}. 
Since $\epsilon\sqrt{\log(M/\epsilon)}\to 0$ as $\epsilon \to 0^+$, it follows that $\epsilon\sqrt{\log N(\mathcal{A}_{\full}, \epsilon)}\to 0$ as $\epsilon \to 0^+$.
\end{proof}

\section{Proof of Theorem~\ref{thm:any-bounds-lower}}

For the any-algorithm lower bound, we use techniques closely related to those used in the proof of Theorem~\ref{thm:geometric-bounds-lower}.
Given an arbitrary subset $\mathcal{A}$, with cardinality $K$, we define auxiliary variables $Y_{\ell,i}$ for $\ell\le m$ and $i\le K$.

\paragraph{Statement of Theorem~\ref{thm:any-bounds-lower}.}
For any algorithm that outputs a subset $\mathcal{A}$ of cardinality $K$, for some absolute constant $C>0$,
\begin{align*}
\E_{\theta}[\regret]
\;\ge\;
\max_{\ell\le m}
\E_{\theta}\!\left[
\max_{a \in r_{\ell}} \mu_a - \mu_{a_{\ell}}
\right]
- C\;\Delta(\mcA)\sqrt{\log m + \log K}.
\end{align*}

\begin{proof}
\textbf{Step 1:}
Fix a pair $(\ell,i)$, define a centered Gaussian process $\left\{Z_a\right\}_{a\in r_{\ell}\cup\{a_i\}}$ where $Z_a:=\mu_a-\mu_{a_i}$. 
Define a non-negative random variable \[
Y_{\ell,i}:=\sup_{a\in r_{\ell}\cup\{a_i\}} Z_a=\sup_{a\in r_{\ell}\cup\{a_i\}} \mu_a - \mu_{a_i}.
\]

\textbf{Step 2:}
Set $\Delta:=\Delta(\mcA)$.
Further, by definition $\E  Z_a^2 = \E (\mu_a-\mu_{a_i})^2=\|a-a_i\|^2_2\leq \Delta^2$.
Using Lemma~\ref{lem:Borell-TIS} on the process $\left\{Z_a\right\}_{a\in r_{\ell}\cup\{a_i\}}$, we have
\begin{equation*}
\Pr\left[\left|Y_{\ell,i} - \E Y_{\ell,i} \right|\geq u\right]\leq 2\exp{\left(-\frac{u^2}{2\Delta^2}\right)}.
\end{equation*}
By union bound, we have
\begin{equation*}
\Pr\left[\max_{\ell\le m, i\le K}\left|Y_{\ell,i} - \E Y_{\ell,i} \right|\geq u\right]\leq 2mK\exp{\left(-\frac{u^2}{2\Delta^2}\right)}.
\end{equation*}
Using Lemma~\ref{lem:integral}, we have for some absolute constant $C>0$
\begin{equation}
\E \max_{\ell\le m, i\le K}\left|Y_{\ell,i} - \E Y_{\ell,i} \right| \leq C\Delta\sqrt{\log mK}.\label{equ:max-deviation-2}
\end{equation}

\textbf{Step 3:}
Now, we connect these auxiliary variables $Y_{\ell,i}$ to regret definition.
Fix $\ell\le m$. Let the cluster contains the optimal action, i.e., $a^*(\theta)\in r_{\ell}$, we have
\begin{align*}
\regret 
&= \max_{a \in \mathcal{A}_{\full}} \mu_a - \max_{a' \in \mathcal{A}} \mu_{a'} \\
&=\max_{a \in r_{\ell}} \mu_a - \max_{i\le K} \mu_{a_i}\\
&=\min_{i\le K} \left\{\max_{a \in r_{\ell}} \mu_a- \mu_{a_i}\right\}\\
&=\min_{i\le K} \left\{\max_{a \in r_{\ell}\cup\{a_i\}} \mu_a- \mu_{a_i}\right\}\\
&= \min_{i\le K} Y_{\ell,i}\\
&= \min_{i\le K} \max_{\ell'\le m} Y_{\ell',i}
\end{align*}
The second equality uses $a^*(\theta)\in r_{\ell}$ and $\mcA=\{a_i\}_{i\le K}$ is action subset.
The fourth equality uses under the event of $a^*(\theta)\in r_{\ell}$, we have
\[
\max_{a\in \mcAF} \mu_a=\max_{a\in r_{\ell}} \mu_a \ge \mu_{a_i}
\]
such that $\max_{a \in r_{\ell}\cup\{a_i\}} \mu_a= \max_{a \in r_{\ell}} \mu_a$.
The fifth equality uses the definition of $Y_{\ell,i}$.
The last uses that under the event of $a^*(\theta)\in r_{\ell}$, we have $\max_{a\in r_{\ell}}\mu_a \ge \max_{a\in r_{\ell'}}\mu_a$ for any $\ell'\le m$, such that
\[
Y_{\ell,i}\ge Y_{\ell',i} \quad\forall \ell'\le m.
\]
Therefore, since $\ell$ is included in $1,\dots,m$, we reach the fact that
\[
Y_{\ell,i}=\max_{\ell'\le m} Y_{\ell',i}. 
\]
Note that this term $\min_{i\le K} \max_{\ell'\le m} Y_{\ell',i}$ is already irrelevant to $\ell$. We conclude that for any $\theta$,
\[
\regret \ge \min_{i\le K} \max_{\ell\le m} Y_{\ell,i}. 
\]

\textbf{Step 4:}
Using $Y_{\ell, i} = \E Y_{\ell, i} + (Y_{\ell, i}- \E Y_{\ell, i})$, we have
\begin{equation}
\begin{aligned}
\min_{i\le K}\max_{\ell \le m} Y_{\ell,i}
&=
\min_{i\le K}\max_{\ell \le m}
\Bigl(
\mathbb{E}Y_{\ell,i} + (Y_{\ell,i}-\mathbb{E}Y_{\ell,i})
\Bigr) \\
&\ge
\min_{i\le K}\max_{\ell \le m}
\Bigl(
\mathbb{E}Y_{\ell,i} - |Y_{\ell,i}-\mathbb{E}Y_{\ell,i}|
\Bigr) \\
&\ge
\min_{i\le K}\max_{\ell \le m}
\Bigl(
\mathbb{E}Y_{\ell,i} -
\max_{i'\le K}\max_{\ell' \le m}
|Y_{\ell',i'}-\mathbb{E}Y_{\ell',i'}|
\Bigr) \\
&=
\min_{i\le K}\max_{\ell \le m}\mathbb{E}Y_{\ell,i}
-
\max_{i' \le K}\max_{\ell' \le m}
|Y_{\ell',i'}-\mathbb{E}Y_{\ell',i'}|.
\end{aligned}
\label{equ:min-max-decomp-2}
\end{equation}
The first step uses the decomposition
\(Y_{\ell,i}=\mathbb{E}Y_{\ell,i}+(Y_{\ell,i}-\mathbb{E}Y_{\ell,i})\).
The second step follows from the inequality \(x \ge x - |x|\), applied pointwise.
For the third step, we use that for all \((\ell,i)\),
\[
|Y_{\ell,i}-\mathbb{E}Y_{\ell,i}|
\le
\max_{i' \le K}\max_{\ell'\le m}
|Y_{\ell',i'}-\mathbb{E}Y_{\ell',i'}|,
\]
so each absolute value can be uniformly upper bounded. In the last step, the
term
\[
\max_{i' \le K}\max_{\ell'\le m}
|Y_{\ell',i'}-\mathbb{E}Y_{\ell',i'}|
\]
does not depend on \((\ell,i)\), so it can be taken outside the inner maximum.

\textbf{Step 5:}
Recall the definition of $Y_{\ell,i}$, fix $\ell$
\begin{equation}
\min_{i\le K}\mathbb{E}Y_{\ell,i} 
= \min_{i\le K}\E\left[ \max_{a\in r_{\ell}\cup\{a_i\}} \mu_a - \mu_{a_i}\right]=\min_{i\le K}\E\left[ \max_{a\in r_{\ell}\cup\{a_i\}} \mu_a \right]\ge \E\left[ \max_{a\in r_{\ell}} \mu_a \right]
=\E\left[ \max_{a\in r_{\ell}} \mu_a-\mu_{a_\ell} \right],
\label{equ:min-EYll-2}
\end{equation}
where the first equality use the definition of $Y_{\ell,i}$. The second uses $\E \mu_{a_i}=0$.
The inequality uses 
\(
\max_{a\in r_{\ell}\cup\{a_i\}} \mu_a  \ge \max_{a\in r_{\ell}} \mu_a
\)
for all $i\le K$.
The last step uses again $\E \mu_{a_\ell}=0$, where $a_{\ell}$ is the representative for cluster $r_{\ell}$.

\textbf{Step 6: Wrap up.}
\begin{align*}
\E[\regret]
&\ge \min_{i\le K}\max_{\ell \le m}\mathbb{E}Y_{\ell,i}
-
\E\max_{i' \le K}\max_{\ell'\le m}
|Y_{\ell',i'}-\mathbb{E}Y_{\ell',i'}|\\
&\ge \max_{\ell \le m}\min_{i\le K}\mathbb{E}Y_{\ell,i}
-
\E\max_{i' \le K}\max_{\ell'\le m}
|Y_{\ell',i'}-\mathbb{E}Y_{\ell',i'}|\\
&\ge \max_{\ell \le m}\E\left[ \max_{a\in r_{\ell}} \mu_a-\mu_{a_i} \right]
-C\Delta\sqrt{\log mK}.
\end{align*}
The first use Equation~\eqref{equ:min-max-decomp-2}.
The second uses the same min-max swap as proved in Equation~\eqref{equ:min-max-reverse}.
The last uses Equations~\eqref{equ:min-EYll-2} and~\eqref{equ:max-deviation-2}.
\end{proof}

\section{Proof of Theorem~\ref{thm:subGaussian-upper-bound}}
\label{sec:subGaussian-bound-proof}

The Gaussian process assumption can be relaxed to a sub-Gaussian one, where the linear structure in Equation~\eqref{equ:mu-define} is not required, and the random process $\{\mu_a\}_{a \in \mathcal{S}}$ satisfies the following increment condition:
\begin{equation*}
\forall u > 0,\;\Pr\left[|\mu_a \!-\! \mu_{a'}| \geq u\right] \leq 2 \exp\left(-\frac{u^2}{2\|a\!-\!a'\|_2^2}\right).
\end{equation*}

We first prove the following lemmas using Talagrand’s $\gamma_2$ functional and the generic-chaining \citep{talagrand2014upper}.

\paragraph{Statement of Lemma~\ref{lem:sub-gaussian-tool}}
Let $T \subset \mathbb{R}^d$ be finite.
Let the random process $\{X_t\}_{t\in T}$ satisfy the increment condition \eqref{equ:increment-define}.
Then, for every $u \ge 1$,
\[
\mathbb{P}\left(
\sup_{s,t \in T} |X_t - X_s|
\le 16\,\gamma_2(T) + 8\,u\,\diam(T)
\right)
\ge 1 - 2 e^{-u^2}.
\]

\begin{proof}
Proof of Lemma~\ref{lem:sub-gaussian-tool}.
Since $T$ is finite, there exists an admissible sequence $(T_n)_{n \ge 0}$ such that
\[
|T_0| = 1, \qquad |T_n| \le 2^{2^n}, \qquad
\sup_{t \in T} \sum_{n \ge 0} 2^{n/2} \mathrm{dist}(t, T_n) = \gamma_2(T),
\]
where $\mathrm{dist}(t, T_n) := \inf_{a \in T_n} \|t-a\|_2$.

For each $t \in T$, choose $\pi_n(t) \in T_n$ such that
\[
\|t - \pi_n(t)\|_2 = \mathrm{dist}(t, T_n).
\]
For $n$ large enough we have $T_n = T$, hence $\pi_n(t) = t$.

Fix $t_0 \in T_0$. Then
\[
\sup_{s,t \in T} |X_t - X_s|
\le 2 \sup_{t \in T} |X_t - X_{t_0}|.
\]

Fix $u \ge 1$ and choose $\kappa$ such that
\[
2^{\kappa/2} \le u < 2^{(\kappa+1)/2}.
\]

For each $t \in T$,
\[
X_t - X_{t_0}
=
\underbrace{(X_{\pi_\kappa(t)} - X_{t_0})}_{\text{coarse part}}
+
\underbrace{\sum_{n>\kappa} \big(X_{\pi_n(t)} - X_{\pi_{n-1}(t)}\big)}_{\text{fine part}}.
\]

\medskip

\noindent
\textbf{Step 1: Coarse part.}

Since $\pi_\kappa(t) \in T_\kappa$,
\[
\sup_{t \in T} |X_{\pi_\kappa(t)} - X_{t_0}|
\le \max_{a \in T_\kappa} |X_a - X_{t_0}|.
\]

Due to the increment condition \eqref{equ:increment-define}, for each $a \in T_\kappa$,
\[
\mathbb{P}\big(|X_a - X_{t_0}| > r\,\diam(T)\big)
\le 2 e^{-r^2/2}.
\]

Using $|T_\kappa| \le 2^{2^\kappa}$ and union bound,
\[
\mathbb{P}\left(
\max_{a \in T_\kappa} |X_a - X_{t_0}| > r\,\diam(T)
\right)
\le 2^{2^\kappa} \cdot 2 e^{-r^2/2}.
\]

Take $r = 4u$. Since $2^\kappa \le u^2$,
\[
2^{2^\kappa} \cdot 2 e^{-8u^2}
\le 2 \exp\big((\log 2)u^2 - 8u^2\big)
\le e^{-u^2}.
\]

Thus with probability at least $1 - e^{-u^2}$,
\[
\sup_{t \in T} |X_{\pi_\kappa(t)} - X_{t_0}|
\le 4u\,\diam(T).
\]

\medskip

\noindent
\textbf{Step 2: Fine part.}

For $n>\kappa$, define the event
\[
E_n :=
\left\{
|X_{\pi_n(t)} - X_{\pi_{n-1}(t)}|
\le 3\cdot 2^{n/2} \|\pi_n(t) - \pi_{n-1}(t)\|_2
\ \text{for all } t \in T
\right\}.
\]

For each $n$, there are at most
\[
|T_n||T_{n-1}| \le 2^{2^n} 2^{2^{n-1}} \le 2^{2^{n+1}}
\]
pairs to check. Duo to the increment condition \eqref{equ:increment-define}, for each pair,
\[
\mathbb{P}\big(|X_a - X_b| > 3\cdot 2^{n/2}\|a-b\|_2\big)
\le 2 e^{-(9/2) \cdot 2^n}.
\]

Hence, by union bound, 
\[
\mathbb{P}(E_n^c)
\le 2^{2^{n+1}} \cdot 2 e^{-(9/2)\cdot 2^n}
= e^{(2\log 2)\cdot 2^n}\cdot 2 e^{-(9/2)\cdot 2^n}
\le 2 e^{((2\log 2)-(9/2))\cdot 2^n} \le 2 e^{-3\cdot 2^n}.
\]

Therefore, using union bound and $2^{\kappa+1}\ge u^2$,
\[
\mathbb{P}\Big(\bigcup_{n>\kappa} E_n^c\Big)
\le \sum_{n>\kappa} 2 e^{-3\cdot 2^n}\le \frac{2e^{-3\cdot 2^{\kappa+1}}}{1-e^{-3}}
\le e^{-u^2}.
\]

On the event $\cap_{n>\kappa} E_n$, for every $t \in T$,
\[
\sum_{n>\kappa} |X_{\pi_n(t)} - X_{\pi_{n-1}(t)}|
\le 3\sum_{n>\kappa} 2^{n/2} \|\pi_n(t) - \pi_{n-1}(t)\|_2.
\]

Using
\[
\|\pi_n(t) - \pi_{n-1}(t)\|_2
\le \mathrm{dist}(t, T_n) + \dist(t,T_{n-1}),
\]
we obtain
\[
\sum_{n>\kappa} 2^{n/2} \|\pi_n(t) - \pi_{n-1}(t)\|_2
\le
\sum_{n>\kappa} 2^{n/2} \mathrm{dist}(t, T_n)
+
\sum_{n>\kappa} 2^{n/2} \dist(t,T_{n-1}).
\]

The second term satisfies
\[
\sum_{n>\kappa} 2^{n/2} \dist(t,T_{n-1})
=
\sqrt{2} \sum_{k\ge \kappa} 2^{k/2} \dist(t,T_k).
\]

Hence
\[
\sum_{n>\kappa} 2^{n/2} \|\pi_n(t) - \pi_{n-1}(t)\|_2
\le (1+\sqrt{2}) \sum_{k\ge 0} 2^{k/2} \dist(t,T_k).
\]

Taking supremum over $t$, and by definition of $\gamma_2(T)$,
\[
\sup_{t \in T} \sum_{n>\kappa} |X_{\pi_n(t)} - X_{\pi_{n-1}(t)}|
\le 3(1+\sqrt{2})\, \gamma_2(T).
\]

\medskip

\noindent
\textbf{Step 3: Wrap up.}

With probability at least $1 - 2 e^{-u^2}$, both bounds hold, and therefore
\[
\sup_{t \in T} |X_t - X_{t_0}|
\le 3(1+\sqrt{2})\, \gamma_2(T) + 4u\,\diam(T).
\]

Thus
\[
\sup_{s,t \in T} |X_t - X_s|
\le 2 \sup_{t \in T} |X_t - X_{t_0}|
\le 16\, \gamma_2(T) + 8u\,\diam(T).
\]
This completes the proof.
\end{proof}

\paragraph{Statement of Lemma~\ref{lem:sub-Gaussian-deviation}:}
Let $\{X_t\}_{t\in T}$ be a sub-Gaussian process satisfying the increment condition \eqref{equ:increment-define}, then there exists an absolute constant $C>0$ that for any $u>0$
\[
\Pr\left(\sup_{s,t \in T} |X_t - X_s|\geq u\;\gamma_2(T)\right)\leq C\exp\left(-u^2/2\right).
\]
\begin{proof}
\textbf{Step 1.}
Let $(T_n)_{n\ge0}$ be an admissible sequence achieving $\gamma_2(T)$, and let
$\pi_n(t)\in T_n$ satisfy $\|t-\pi_n(t)\|_2=\dist(t,T_n)$.
Since
\[
  \sup_{s,t\in T}|X_t-X_s|
  \le
  2\sup_{t\in T}|X_t-X_{t_0}|,
\]
it suffices to bound the latter. Fix $t\in T$ and write
\[
  X_t-X_{t_0}
  =
  \sum_{n\ge1}
  \bigl(X_{\pi_n(t)}-X_{\pi_{n-1}(t)}\bigr).
\]

\textbf{Step 2.}
Set $\beta:=2\sqrt{\log 2}$, and for each $n\ge1$ define the event
\[
  E_n:=
  \Bigl\{
    |X_a-X_b|
    \le
    (u+\beta)\,2^{n/2}\|a-b\|_2
    \quad
    \forall\,a\in T_n,\ b\in T_{n-1}
  \Bigr\}.
\]
Since $|T_n||T_{n-1}|\le 2^{2^{n+1}}$, the increment
condition~\eqref{equ:increment-define} and a union bound give
\[
  \Pr(E_n^c)
  \le
  2^{2^{n+1}}\cdot 2e^{-(u+\beta)^2 2^n/2}
  \le
  2e^{-u^2 2^n/2}.
\]
Summing over $n\ge1$,
\[
  \Pr\!\Big(\bigcup_{n\ge1}E_n^c\Big)
  \le
  2\sum_{n\ge1}e^{-u^2 2^n/2}
  \le
  C e^{-u^2/2}
\]
for an absolute constant $C>0$.

\textbf{Step 3.}
On the event $\bigcap_{n\ge1}E_n$, for every $t\in T$,
\[
  |X_t-X_{t_0}|
  \le
  (u+\beta)
  \sum_{n\ge1}
  2^{n/2}
  \|\pi_n(t)-\pi_{n-1}(t)\|_2.
\]
Using $\|\pi_n(t)-\pi_{n-1}(t)\|_2\le\dist(t,T_n)+\dist(t,T_{n-1})$
and the definition of $\gamma_2(T)$,
\[
  \sum_{n\ge1}
  2^{n/2}
  \|\pi_n(t)-\pi_{n-1}(t)\|_2
  \lesssim
  \gamma_2(T).
\]
Hence $\sup_{s,t\in T}|X_t-X_s|\lesssim(u+\beta)\,\gamma_2(T)$ with
probability at least $1-Ce^{-u^2/2}$.
Replacing $u$ by $u/C'$ for a suitable absolute constant $C'$ and
absorbing $\beta$ into $C'$ yields the stated bound.
\end{proof}

\paragraph{Statement of Theorem~\ref{thm:subGaussian-upper-bound}:}
Let $\mathcal{A}$ be output of Algorithm~\ref{alg:smallvertices}. 
Let $\{\mu_a\}_{a\in\mathcal{A}_{\full}}$ be a sub-Gaussian process satisfying Equation~\eqref{equ:increment-define}. 
Consider a partition \( \mathcal{R} := \{r_{\ell}\}_{\ell \leq m} \) of the full action space, where $m\ge 2$.
Then, for an absolute constant $C>0$
\begin{equation*}
\begin{split}
\E _{\theta,\mathcal{A}}[\textnormal{Regret}] 
\leq & 
16\,\max_{\ell\le m} \, \gamma_2(r_{\ell}) + 29\,\epsilon \sqrt{\log(m)} + C\cdot\left(\E _{q}\left[(1-q(r))^{2K}\right]\cdot\gamma_2^2(\mathcal {A}_{\full})\right)^{1/2}.
\end{split}
\end{equation*}

\begin{proof}
\textbf{Step 1: Auxiliary variables.}
If \( r_{\ell} \cap \mathcal{A} \neq \emptyset \), define \( a_{\ell} \in r_{\ell} \cap \mathcal{A} \). 
If $r_{\ell}\cap\mathcal{A}\!=\!\emptyset$, choose an arbitrary point $a_{\ell}\in r_{\ell}$ as the representative.
The set $\mathcal{A}':=\{a_{\ell}\}_{\ell\leq m}$ forms a reference subset of Definition~\ref{def:ref-subset}. 

For each $\ell\leq m$, define a random variable $$Y_{\ell}:=\sup_{a\in r_{\ell}} \mu_a-\mu_{a_{\ell}}.$$

\textbf{Step 2:}
By Lemma~\ref{lem:sub-gaussian-tool}, we have
for every $u \ge 1$,
\[
\mathbb{P}\left(
\sup_{s,t \in r_{\ell}} |\mu_t - \mu_s|
\le 16\,\gamma_2(r_{\ell}) + 8\,u\,\diam(r_{\ell})
\right)
\ge 1 - 2 e^{-u^2}.
\]
A simple rearrangement gives
for every $\nu \ge 0$,
\[
\mathbb{P}\left(
\sup_{s,t \in r_{\ell}} |\mu_t - \mu_s|
> 16\,\gamma_2(r_{\ell})+ 8\,\diam(r_{\ell}) + \nu
\right)
\le 2\exp\left(-\frac{\nu^2}{8^2\diam(r_{\ell})^2}\right).
\]

Recall that
\[
Y_{\ell}
=\sup_{a\in r_{\ell}} \mu_a-\mu_{a_{\ell}}
\;\le\;
\sup_{s,t \in r_{\ell}} |\mu_t - \mu_s|,
\]
and recall the definition
\[
\epsilon:=\max_{\ell\le m} \diam(r_{\ell}).
\]
Moreover, set
\[
M:=\max_{\ell\le m} \, 16\,\gamma_2(r_{\ell})+ 8\,\epsilon .
\]

By a union bound,
\begin{equation}
\mathbb{P}\!\left(
\max_{\ell\le m} Y_{\ell}
> M + \nu
\right)
\le 2m\exp\!\left(-\frac{\nu^2}{8^2 \epsilon^2}\right).  
\label{equ:max-Y-deviaton}
\end{equation}

For any \(\nu \ge 0\), using Equation~\eqref{equ:max-Y-deviaton}, 
\begin{align*}
\E \max_{\ell \le m} Y_\ell
&\le M + \nu + \int_{\nu}^{\infty}
\mathbb{P}\!\left(\max_{\ell \le m} Y_\ell > M + t\right)\,dt \\
&\le M + \nu + \int_{\nu}^{\infty}
2m \exp\!\left(-\frac{t^2}{8^2 \epsilon^2}\right)\,dt.
\end{align*}

Choose
\[
\nu = 8\,\epsilon \sqrt{\log(2m)}.
\]
Using the Gaussian tail bound
$\int_{\nu}^{\infty} e^{-t^2/a^2}\,dt \le \frac{a^2}{2\nu} e^{-\nu^2/a^2}$
with $a = 8\epsilon$, and noting that $e^{-\nu^2/8^2\epsilon^2} = \frac{1}{2m}$, we obtain
\[
\int_{\nu}^{\infty}
2m \exp\!\left(-\frac{t^2}{8^2 \epsilon^2}\right)\,dt
\le 2m \cdot \frac{8^2\epsilon^2}{2\nu} \cdot \frac{1}{2m}
= \frac{8^2 \epsilon^2}{2\nu}.
\]



Therefore, for $m\ge 2$,
\begin{equation}
\begin{aligned}
\E \max_{\ell \le m} Y_\ell
\le& M + 8\,\epsilon \sqrt{\log(2m)}
+ \frac{4\,\epsilon}{\sqrt{\log(2m)}}\\
\le& M + 12\,\epsilon \sqrt{\log(2m)}\\
= & \max_{\ell\le m} \, 16\,\gamma_2(r_{\ell})+ 8\,\epsilon + 12\,\epsilon \sqrt{\log(2m)}\\
\le& 16\,\max_{\ell\le m} \, \gamma_2(r_{\ell}) + 29\,\epsilon \sqrt{\log(m)}.
\end{aligned}
\label{equ:sub-gaussian-Y}
\end{equation}

\textbf{Step 3:}
Let $r\cap\mathcal{A}=\emptyset$ and $a^*(\theta)\in r$. Since $\mathcal{A}\subseteq\mathcal{A}_{\full}$, the regret is upper bounded by
\begin{equation}
\regret=\max_{a\in\mathcal{A}_{\full}} \mu_a - \max_{a'\in\mathcal{A}} \mu_{a'} \le \sup_{a,a'\in\mathcal{A}_{\full}} |\mu_a-\mu_{a'}|.
\label{equ:sub-gaussian-worstcasebound}
\end{equation}

\textbf{Step 4:}
Let $X:=\sup_{a,a'\in\mathcal S} |\mu_a-\mu_{a'}|$, and $\sigma:=\gamma_2(\mathcal S)$.
By Lemma~\ref{lem:sub-Gaussian-deviation}, there is an absolute constant $C>0$ that
\[
\Pr(X\ge u)\le C\exp\left(-\frac{u^2}{2\sigma^2}\right).
\]

Using the tail integration formula,
\begin{align*}
\mathbb E[X^2]
&=
\int_0^\infty \Pr(X^2\ge t)\,dt\\
&=
\int_0^\infty 2u\,\Pr(X\ge u)\,du\\
&\le \int_0^\infty
2u \min\left\{1,\,
C e^{-u^2/(2\sigma^2)}
\right\}du\\
&\le 
\int_0^{u_0} 2u\,du
+
2C\int_{u_0}^\infty
u e^{-u^2/(2\sigma^2)}du \qquad\text{set $u_0:=\sigma\sqrt{2\log C}$}\\
&\le
2\sigma^2\log C
+
2\sigma^2,
\end{align*}
where the second equality uses the change of variables. The first inequality uses Lemma~\ref{lem:sub-Gaussian-deviation}, and $\Pr(X\ge u)\le 1$.
The second inequality splits the integral at \(u_0=\sigma\sqrt{2\log C}\).
For the last inequality, the first term satisfies
\[
\int_0^{u_0}2u\,du
=
u_0^2
=
2\sigma^2\log C;
\]
The second term satisfies
\[
2C\int_{u_0}^\infty
u e^{-u^2/(2\sigma^2)}du
=
2C\sigma^2 e^{-u_0^2/(2\sigma^2)}
=
2\sigma^2.
\]

Therefore, set $C'=1+\log C$, we have
\begin{equation}
\mathbb E\!\left[
\left(\sup_{a,a'\in\mathcal S}|\mu_a-\mu_{a'}|\right)^2
\right]
\le
C'\; 2\gamma_2^2(\mathcal S). 
\label{equ:sub-gaussian-2moment}
\end{equation}

\textbf{Step 5:}
Following the same reasoning used in the proof of Theorem~\ref{thm:alg-bound-upper}:
\begin{align*}
\E [\regret] = & \sum_{r\in\mathcal{R}} \Pr[r \cap\mathcal{A}\neq\emptyset,a^*(\theta)\in r ]\cdot\E \left[\regret \Big| r \cap\mathcal{A}\neq\emptyset,a^*(\theta)\in r  \right]\\
& + \sum_{r\in\mathcal{R}} \Pr[r \cap\mathcal{A}=\emptyset,a^*(\theta)\in r ]\cdot\E \left[\regret \Big| r \cap\mathcal{A}=\emptyset,a^*(\theta)\in r  \right]\\
\leq & \sum_{\ell\leq m} q(r_{\ell})\cdot \left(\E \left[\max_{\ell\leq m} Y_{\ell} \Big|a^*(\theta)\in r_{\ell} \right] +  (1-q(r_{\ell}))^K\cdot \E \left[\sup_{a,a'\in\mathcal{A}_{\full}} |\mu_a-\mu_{a'}|\Big|a^*(\theta)\in r_{\ell}\right] \right) \\
\leq & \E \left[\max_{\ell\leq m} Y_{\ell}\right] + \left(\E _{q}\left[(1-q(r))^{2K}\right]\cdot \E _{\theta}\left[\left(\sup_{a,a'\in\mathcal{A}_{\full}} |\mu_a-\mu_{a'}|\right)^2\right]\right)^{1/2} \\
\leq & 16\,\max_{\ell\le m} \, \gamma_2(r_{\ell}) + 29\,\epsilon \sqrt{\log(m)} + \left(\E _{q}\left[(1-q(r))^{2K}\right]\cdot C''\gamma_2^2(\mathcal {A}_{\full})\right)^{1/2},
\end{align*}
The first inequality uses \eqref{equ:sub-gaussian-worstcasebound}, and the same reasoning in the proof of Theorem~\ref{thm:alg-bound-upper}.
The second inequality follows from the Cauchy--Schwarz and Jensen inequalities.
The last inequality uses Equation~\eqref{equ:sub-gaussian-Y}, and \eqref{equ:sub-gaussian-2moment}, where $C''>0$ is an absolute constant.
\end{proof}

\section{Properties of Gaussian width}
\label{app:gaussian-width}
\noindent
Given a set $\mathcal{S}\in\mathbb{R}^n$, the term $\E [\max_{a\in\mathcal{S}}\mu_a]$ where $\theta\sim\mathcal{N}(0,I)$ is called Gaussian (mean) width. 
\paragraph{Property 1: $\E \left[\max_{a\in\mathcal{S}}\langle a,\theta\rangle\right]=\E \left[\max_{a'\in-\mathcal{S}}\langle a',\theta\rangle\right]$.}
\begin{align*}
\E \left[\max_{a\in\mathcal{S}}\langle a,\theta\rangle\right]=\E \left[\max_{a\in\mathcal{S}}\langle a,-\theta\rangle\right]=\E \left[\max_{a\in\mathcal{S}}\langle -a,\theta\rangle\right] =\E \left[\max_{a'\in-\mathcal{S}}\langle a',\theta\rangle\right],
\end{align*}
where the first equality uses $-\theta$ and $\theta$ are identically distributed. The third equality uses for any $a\in\mathcal{S}$, it holds that $-a\in-\mathcal{S}$.
\paragraph{Property 2: $\E \left[\max_{a\in\mathcal{S}}\langle a,\theta\rangle\right]\leq \frac{1}{2}\E \left[\max_{a,a'\in\mathcal{S}}\langle a-a',\theta\rangle\right]$.} Let $a^*(-\mathcal{S},\theta)$ denote the optimal action in $\mathcal{S}$ for bandit instance $\theta$.
\begin{align*}
2\cdot \E \left[\max_{a\in\mathcal{S}}\langle a,\theta\rangle\right]=&\E \left[\max_{a\in\mathcal{S}}\langle a,\theta\rangle\right]+\E \left[\max_{a'\in-\mathcal{S}}\langle a',\theta\rangle\right]\\
=&\E \left[\langle a^*(\mathcal{S},\theta),\;\theta\rangle\right]+\E \left[\langle a^*(-\mathcal{S},\theta),\;\theta\rangle\right]\\
=& \E \left[\langle a^*(\mathcal{S},\theta)+a^*(-\mathcal{S},\theta),\;\theta\rangle\right]\\
\leq & \E \left[\max_{a,a'\in\mathcal{S}}\langle a-a',\theta\rangle\right],
\end{align*}
where the first equality uses Property 1, the second equality uses the definition of $a^*(-\mathcal{S},\theta)$. The third equality uses linearity of expectation. The inequality uses that the vector $a^*(\mathcal{S},\theta)+a^*(-\mathcal{S},\theta)$ belongs to the set of vectors $\{a-a':a,a'\in\mathcal{S}\}$. In fact, one can prove the equality that $\E \left[\max_{a\in\mathcal{S}}\langle a,\theta\rangle\right]=\frac{1}{2}\E \left[\max_{a,a'\in\mathcal{S}}\langle a-a',\theta\rangle\right]$, but we only need inequality to prove the next claim.
\paragraph{Property 3: $\E \left[\max_{a\in\mathcal{S}}\mu_a\right]\leq \frac{\diam(\mathcal{S})}{2}\cdot\sqrt{n}$.}
\begin{align*}
\E \left[\max_{a\in\mathcal{S}}\mu_a\right]=&\E \left[\max_{a\in\mathcal{S}}\langle a,\theta\rangle\right]\\
\leq &\frac{1}{2}\E \left[\max_{a,a'\in\mathcal{S}}\langle a-a',\theta\rangle\right] \\
\leq& \frac{1}{2}\E \max_{a,a'\in\mathcal{S}}\|\theta\|_2\|a-a'\|_2 \\
\leq&\frac{1}{2}\E \diam(\mathcal{S}) \|\theta\|_2 \leq \frac{\diam(\mathcal{S})}{2}\cdot\sqrt{n},
\end{align*}
where the first equality uses the definition of $\mu_a$. The first inequality uses Property 2. The second inequality uses Cauchy-Schwarz inequality. The third inequality uses the definition of $\diam(\cdot)$. The last inequality uses $\E \|\theta\|_2\leq \sqrt{n}$.

\section{Bounds of expected maximum of Gaussian}
\label{app:bound-maxgaussian}
Let $X_1,\dots,X_N$ be $N$ random Gaussian variables (no necessarily independent) with zero mean and variance of marginals smaller than $\sigma^2$, then
\begin{equation*}
\E \left[\max_{i=1,\dots,N} X_i\right]\leq \sigma\sqrt{2\log N}.
\end{equation*}
\begin{proof}
for any $\delta>0$,
\begin{align*}
\E \left[\max_{i=1,\dots,N} X_i\right] =& \frac{1}{\delta} \E \left[\log \exp(\delta\max_{i=1,\dots,N} X_i)\right]
\leq \frac{1}{\delta}\log \E \left[ \exp(\delta\max_{i=1,\dots,N} X_i)\right]\\
= & \frac{1}{\delta}\log \E \left[ \max_{i=1,\dots,N}  \exp(\delta X_i)\right]
\leq  \frac{1}{\delta}\log \sum_{i=1}^N \E \left[\exp(\delta X_i)\right]\\
\leq & \frac{1}{\delta}\log \sum_{i=1}^N \exp(\sigma^2\delta^2/2) = \frac{\log N}{\delta}+\frac{\sigma^2\delta}{2},
\end{align*}
where the first inequality uses Jensen's inequality. Taking $\delta:=\sqrt{2(\log N)/\sigma^2}$ yields the results. 
\end{proof}

\noindent Let $X_1,\dots,X_N$ be i.i.d. $\mathcal{N}(0,\sigma^2)$ random variables, then according to \citep{kamath2015bounds}:
\begin{equation*}
\E _{\theta}\left[\max_{i=1,\dots,N} X_i\right]\geq \frac{\sigma\sqrt{\log N}}{\sqrt{\pi \log 2}}.
\end{equation*}

\section{The effect of clustering structure on regret}
\label{app:diameter}

\begin{figure*}[t]
\centering
\includegraphics[width=\linewidth]{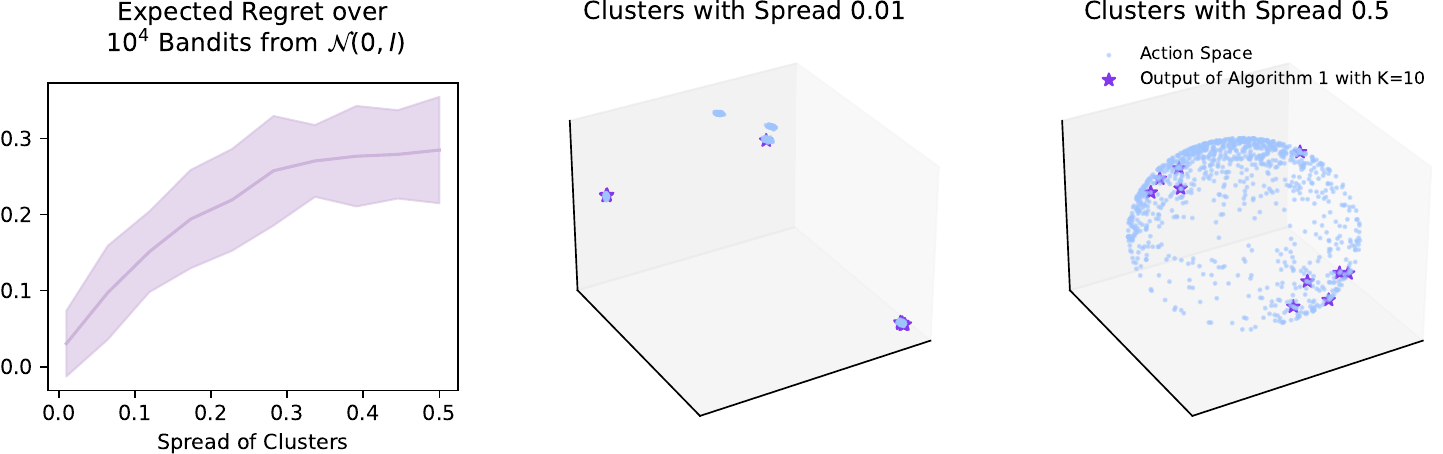}
\caption{Illustration of clustered action spaces on unit sphere in $\mathbb{R}^3$ and the effect of cluster diameters on regret.
Five clusters are formed by generating 5 fixed center points, with 200 points sampled around each using Gaussian noise (spread controls the variance).
Bandits are drawn from $\mathcal{N}(0, I)$.
The left subplot shows the mean $\pm$ standard deviation of the expected regret (over 30 trials) as the spread varies from 0.01 to 0.5, using $10^4$ additional bandits.
The middle and right subplots show example action spaces (blue dots) for spread values 0.01 and 0.5, with representative actions (purple stars) selected by Algorithm~\ref{alg:smallvertices} with $K = 10$.}
\label{fig:cluster}
\end{figure*}

In Figure~\ref{fig:cluster}, we study the effect of the cluster diameters (controlled by a spread parameter)  on regret.
We structure the clustered action space:
Five center points are fixed on the unit sphere in $\mathbb{R}^3$, and around each center, 200 points are sampled to form five clusters.
Each point is obtained by adding Gaussian noise (mean zero, standard deviation equal to the spread parameter) to the center direction, followed by projection back onto the unit sphere.
Bandits are sampled from a 3-dimensional standard Gaussian distribution, i.e., $\theta\sim\mathcal{N}(0,I)$.
The left subplot shows the expected regret of Algorithm~\ref{alg:smallvertices} with $K = 10$, computed using $10^4$ additional bandits, as the spread varies from 0.01 to 0.5. The curves and error shade represent the mean $\pm$ one standard deviation of expected regret over 30 repetitions. The middle and right subplots display example action spaces for spread values of 0.01 and 0.5, with representative actions (purple stars) selected by Algorithm~\ref{alg:smallvertices} with $K=10$.

\section{Varying-dependence actions with RBF/Gibbs kernels}
\label{sec:gibbs}

We study the effect of varying action dependence using a Gaussian process with a kernel. To control the degree of dependence, we use stationary RBF kernel and non-stationary Gibbs kernel \citep{williams2006gaussian}.
\begin{equation}
\begin{split}
\underset{\text{RBF}}{k}(a, a')=&\exp\left( -\frac{\|a - a'\|^2}{2 l^2}\right),\\
\underset{\text{Gibbs}}{k}(a, a')=&\sqrt{\frac{2 \, l(a)l(a')}{l(a)^2 \!+\! l(a')^2}}\exp\left( \!-\!\frac{\|a - a'\|^2}{l(a)^2 \!+\! l(a')^2} \right),
\end{split}
\label{equ:kernel-define-app}
\end{equation}
where $l$ is a length-scale parameter and $l(a):=0.1 + 0.9\cdot\exp(-\|a\|^2)$ is a location-dependent length-scale function. 
Both of them control the dependence between actions.
But, unlike the stationary RBF kernels, the Gibbs kernel allows the correlation to depend not only on the distance between actions, but also on their locations.
When $l(a)=l$ is a constant, the Gibbs kernel reduces to the RBF kernel.

\textbf{Sampling Outcome Functions from a Kernel:}
We first construct the kernel matrix $\mathbf{K}$, where each entry is given by $\mathbf{K}_{a,a'} = k(a,a')$, for $a,a'\in\mathcal{A}_{\full}$, depending on the choice of kernel.
We then sample a Gaussian vector (a Gaussian process function evaluated at a finite set of input points) $f \sim \mathcal{N}(0, \mathbf{K})$.
Under either kernels defined in \eqref{equ:kernel-define-app}, the variance of the function value $f(a)$ is one for all $a \in \mathcal{A}_{\full}$.
In this way, we sample functions from a RKHS function class; See \citep[Theorem~4.12]{kanagawa2018gaussian}.

\begin{figure}[t]
\centering
\includegraphics[width=0.4\textwidth]{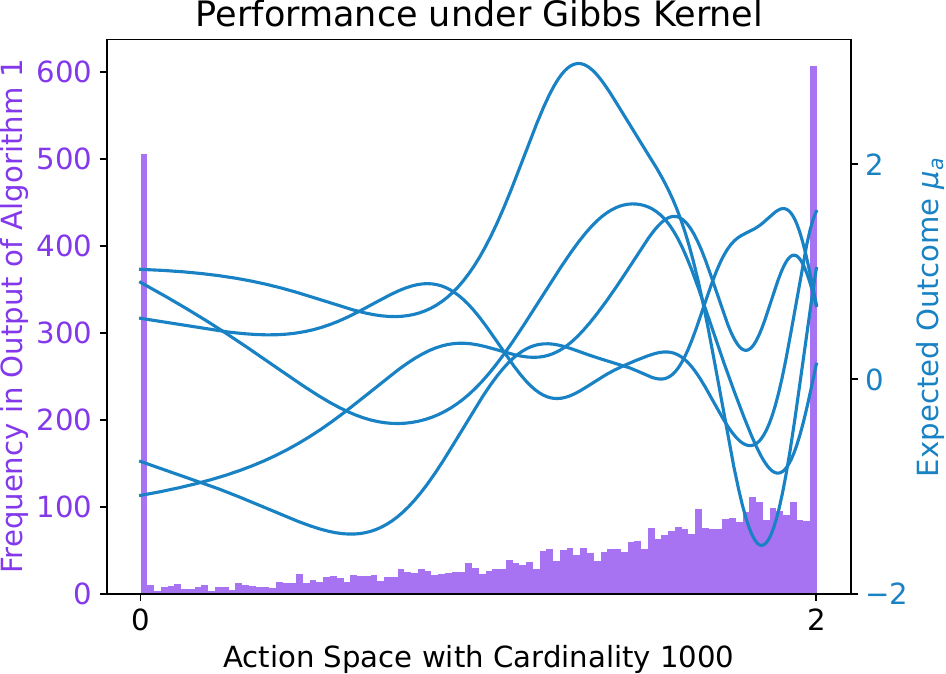}
\caption{Experiments with outcome functions sampled from RBF/Gibbs kernels in \eqref{equ:kernel-define-app}.
Sampled outcome functions from Gibbs kernel over fixed 1000 grid points in $[0,2]$ (blue curves, right y-axis). The histogram (purple bars, left y-axis) shows action selection frequencies by Algorithm~\ref{alg:smallvertices} with $K=5000$, favoring regions with rougher functions and edge points.}
\label{fig:Gibbs}
\end{figure}

To study the effect of varying dependence on the output of Algorithm~\ref{alg:smallvertices}, we consider a fixed action space consisting of 1000 grid points in the interval $[0, 2]$, using Gibbs kernel in \eqref{equ:kernel-define-app} to sample outcome functions.
To simplify computations, we marginalize a Gaussian process defined by the kernel over the grid.
Figure~\ref{fig:Gibbs} provides examples of sampled outcome functions (blue curves, with the y-axis on the right-hand side), which become smoother as the actions approach the left end of the interval---indicating stronger correlations among function values in that region.
We run Algorithm~\ref{alg:smallvertices} on this action space with $K = 5000$ to select actions and record the frequency of each action being selected.
The resulting histogram (purple bars, with the y-axis on the left) reflects the importance measure $q$, highlighting that Algorithm~\ref{alg:smallvertices} tends to select more actions from regions where the outcome functions are rougher---i.e., where action outcomes are less correlated and their features $\Phi(a)$ are farther apart.
Another interesting aspect of this subplot is the two high bars at the edges. Recall that the actual action space consists of feature vectors $\Phi(a)$ for $a\in\mathcal{A}_{ful}$. For actions indexed closer to 0, their feature vectors become more densely packed compared to those indexed closer to 2, resulting in more correlated outcomes. The two actions at the edges, indexed by 0 and 2, correspond to the two farthest points in the actual feature space.

\vskip 0.2in
\bibliographystyle{abbrvnat}
\bibliography{ref}

\end{document}